\RequirePackage{fix-cm}
\documentclass[smallextended]{svjour3}       
\smartqed  
\usepackage{graphicx}
\usepackage{booktabs}
\usepackage{url}
\usepackage{cite}
\usepackage{times}
\usepackage{epsfig}
\usepackage{graphicx}
\usepackage{amsmath}
\usepackage{amssymb}
\usepackage{caption}
\usepackage{adjustbox}
\usepackage{algorithm}
\usepackage{algpseudocode}
\begin{document}

\title{GeoStat Representations of Time Series for Fast Classification
}


\author{Robert J. Ravier \\ Mohammadreza Soltani \\  Miguel Simoes \\ Denis Garagic \\ Vahid Tarokh
}


\institute{R. Ravier \and M. Soltani \and V. Tarokh\at
              Department of ECE \\
              Duke University\\
              \email{\{robert.ravier,mohammadreza.soltani,vahid.tarokh\}@duke.edu}           
           \and
            M. Simoes \at
              Department of EE \\
              KU Leuven\\
              \email{miguel.alfaiatesimoes@kuleuven.be}
            \and 
            D. Garagic \at
              Sarcos Robotics \\
              \email{d.garagic@sarcos.com}
            \and
}

\date{Received: date / Accepted: date}

\maketitle

\begin{abstract}
Recent advances in time series classification have largely focused on methods that either employ deep learning or utilize other machine learning models for feature extraction. Though successful, their power often comes at the requirement of computational complexity. In this paper, we introduce GeoStat representations for time series. GeoStat representations are based off of a generalization of recent methods for trajectory classification, and summarize the information of a time series in terms of comprehensive statistics of (possibly windowed) distributions of easy to compute differential geometric quantities, requiring no dynamic time warping. The features used are intuitive and require minimal parameter tuning. We perform an exhaustive evaluation of GeoStat on a number of real datasets, showing that simple KNN and SVM classifiers trained on these representations exhibit surprising performance relative to modern single model methods requiring significant computational power, achieving state of the art results in many cases. In particular, we show that this methodology achieves good performance on a challenging dataset involving the classification of fishing vessels, where our methods achieve good performance relative to the state of the art despite only having access to approximately two percent of the dataset used in training and evaluating this state of the art.
\end{abstract}

\section{Introduction}\label{sec:intro}
Time series analysis has long been an important focus of quantitative research, largely due to the ubiquitous presence of time series data in a wide variety of applications such as finance, \cite{kumar2002clustering,taylor2008modelling}, medicine \cite{wismuller2002cluster}, illegal fishing \cite{kroodsma2018tracking,boerder2018global}, among others, too many to list here. Contemporary work has focused on utilizing increased computational abilities to move past ARIMA models \cite{box2015time} that, though expressive, heavily depend on certain assumptions that may not hold in practice.

Our focus is time series classification. This subfield has a rich body of work on methodology, and a (relative to other fields) large body of work on testing said methodology, with researchers taking care to validate results in order to determine the state of progress in said field \cite{bagnall2017great}. State of the art methods contemporary to this work primarily consist of ensemble methods \cite{baydogan2013bag,bagnall2015time, lines2016hive, schafer2017fast, fawaz2019inceptiontime, shifaz2020ts}. These methods combine simpler classifiers, such as those based on distances \cite{berndt1994using}, shapelets \cite{ye2009time,hills2014classification}, coefficients of autoregressive models \cite{corduas2008time}, and other features to develop methods that play off the successes and weaknesses of both the features and simpler classifiers. Though accurate, ensemble models can require significant computational resources. TS-CHIEF, one such model, was shown to require multiple days of training in order to evaluate performance on the UCR/UEA 2015 Archive \cite{ucr2018website, dau2019ucr}; this is still a significantly shorter runtime than HIVE-COTE \cite{shifaz2020ts}.

Outside of ensemble models, researchers developing novel classification methods have primarily turned to deep learning based methods, playing off of its success in other fields \cite{cui2016multi, le2016data, wang2017time, zhao2017convolutional,ma2019learning}. Many of these methods were analyzed in \cite{fawaz2019deep}, where it was observed that some of them (in particular, ResNet \cite{he2016deep}) performed extremely well relative to ensemble models. Furthermore, the authors showed empirically that class activation maps \cite{zhou2016learning, song2020representation} could be used in tandem with networks to identify influential regions used in making a particular classification decision. Deep learning classifiers are generally less computationally intensive to train than ensemble models, but still require a fair amount of time: it took approximately 100 GPU days in order to train each of the nine models used in \cite{he2016deep} a total of ten times each.

\begin{figure}[t]
\begin{center}
\includegraphics[scale=.4]{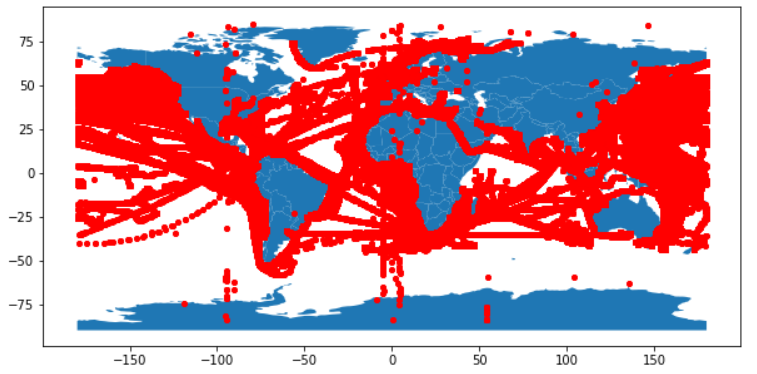}
\end{center}
\caption{Spatial range of one hundred vessels in the Global Fishing Watch dataset \cite{GFWData}. Red indicates a coordinate reached by one of the vessels. Note that there are some obvious errors, e.g. some vessels have coordinates on land.}\label{fig:spatialRange}
\end{figure}

Regardless of the exact type of proposed method, one method remains at the foundation of time series classification: nearest-neighbor Euclidean distance after dynamic time warping (NN-DTW) \cite{jeong2011weighted, bagnall2014experimental,  bagnall2017great}. Highly regarded for its simplicity and ease of implementation, it has long been the baseline to which any novel time series classifier should be compared. Using this classifier as a baseline makes a seemingly innocent but crucial assumption on a given dataset: Euclidean distance after dynamic time warping is a meaningful way to separate time series. Perhaps surprisingly to some readers, this is not always the case. A meaningful real-world counterexample to this idea can be found within a publicly available dataset of naval vessel trajectories over the course of four years provided by Global Fishing Watch \cite{GFWData, GFWWebsite} The trajectories within this dataset consist of twelve separate classes, each of which is a distinct type of vessel. Vessels of the same class can be on extremely different regions of the Earth, suggesting that Euclidean (or any spatial) distance is not a reasonable metric. Further, vessels can have highly different patterns of activity (both in terms of time of movement as well as type of movement), suggesting that dynamic time warping itself may not be meaningful. We defer further dataset details to later sections, but to illustrate the complexity of the problem, we show in Figure~\ref{fig:spatialRange} the spatial range of one hundred trajectories, less than a tenth of those present in the dataset.

Nevertheless, as vessels of the same class do not necessarily correspond to the same region, NN-DTW is not a reasonable classifier to use. The best known classification model for this particular problem is a convolutional neural network with millions of parameters \cite{kroodsma2018tracking}, but this model was trained and tested on a large amount of commercial data not publicly available; the data available in \cite{GFWData, GFWWebsite} is less than two percent of that used in \cite{kroodsma2018tracking}. It is natural to ask, especially given the relatively limited size of the publicly available portion of the data, if a simpler model could still do reasonably well.

In light of the above, there are two questions motivating this paper. First, given the preceding paragraph, we ask whether it is possible to develop a simple model for time series classification that does not require the use of distances between individual points or dynamic time warping. Such a model would allow for a straightforward baseline to benchmark other newly proposed methods with fewer assumptions on the underlying data. Second, we wish to investigate a question stemming from an interesting point made in \cite{he2016deep}. Many in the time series classification community have seemingly ignored whether such problems could be solved by pure feature learning algorithms. This was partially addressed in \cite{lubba2019catch22}, which showed it was possible to learn a small number of features that yielded good classification performance on a large number of datasets. We take this one step further by asking whether easy to compute, mathematically well-understood features exist, without employing any specific algorithm for learning them. We answer both questions we posed in the affirmative, with a surprising twist: all features we use arise from differential geometry.

In this paper, we propose the GeoStat representation for time series data in terms of summary statistics of distributions of well-understood differential geometric features, ignoring many oft-used quantities in time series analysis (e.g. autocorrelation and Fourier transforms). All quantities used are simple to compute in both univariate and multivariate settings, and only require assuming mild smoothness of the time series of interest, which can be easily achieved via standard smoothing methods. We show through extensive empirical evaluation that using (potentially multi-windowed versions of) this representation to train standard $k$NN and SVM classifiers \cite{hastie2009elements} results in surprisingly good classification rates. Specifically, these classifiers have performance that is generally competitive with deep learning methods, and often state of the art; this methodology is, to the best of our knowledge, is the first to achieve perfect classification rates on average for a number of datasets in the oft-used UCR Time Series Classification Archive benchmark \cite{ucr2018website, dau2019ucr}. Our goal, however, is not to claim that our methodology is state of the art: in many cases, it is not. Rather, our goal is to illustrate that this method yields a powerful baseline for time series classification, suggesting that the representation proposed here could be used in more sophisticated methods (e.g. deep learning and ensemble methods) for working with time series.

The paper is outlined as follows. In Section 2 preliminaries and prior work related to our paper. In Section 3, we detail the general pipeline to obtain  features and statistics needed we use for classification. In Section 4, we present results on both the aforementioned vessel trajectory problem as well as on the UCR 2018 archive \cite{dau2019ucr}, demonstrating the effectiveness of our representation and proposed baseline. In Section 5, we evaluate the role of different parameters and features used, showing in part that in many cases we can accurately classify time series using the proposed representation on specific windows in time. We finish with concluding remarks. Additional experiments and details that do not fit within the page constraints will be deferred to the Appendix.
\section{Preliminaries}\label{sec:prelim}
\subsection{Definitions and Notation}

A general (finite) time series is a sequence of points ${\bf{x}} = \{x_{t_{k}}\}_{k=0}^{n}.$ We assume that $x_{t_{k}}$ is a point in $d$-dimensional Euclidean space. Though not all time series we work with are equally spaced in time, i.e. $t_{k+1}-t_{k}$ is not necessarily constant in $k,$ our methods will require this, which can easily be achieved via interpolation. We will specifically use linear interpolation: the linear interpolant $F_{\bf{x}}(t)$ of $\bf{x}$ is defined by, for $t_{0} \leq t_{i} \leq t \leq t_{i+1} \leq t_{n},$

\begin{equation}
F_{\bf{x}}(t) = \frac{t-t_{i}}{t_{i+1}-t_{i}} x_{t_{i}} + \frac{t-t_{i+1}}{t_{i+1}-t_{i}}x_{t_{i+1}}.
\end{equation}\label{eqn:linInterp}

\noindent Our use of linear interpolation is deliberate: it ensures that time series of physical quantities are still physically meaningful (e.g. interpolants of positive quantities remain positive). 

Without further assumptions on $\bf{x},$ it is possible that the interpolated time series will have sharp changes. Many of the quantities of interest for the representation we define in Section 3 require computing derivatives, which do not behave well with sharp changes. To get around this, we employ a simple smoothing operation. Namely, given a time series $\bf{x},$ we define the \emph{Laplacian smoothing} operation $L$ by \\
$L(\bf{x}) = \bf{\tilde{x}},$ where ${\bf{\tilde{x}}} = \{\tilde{x}_{t_{k}}\}_{k=0}^{n}$ is the time series defined by 
$\tilde{x}_{t_{0}} = x_{t_{0}}, \tilde{x}_{t_{n}} = x_{t_{n}},$ and for $1 \leq k \leq n-1,$
\begin{equation}
\tilde{x}_{t_{k}} = \frac{1}{2}(x_{t_{k+1}}-x_{t_{k}}) + \frac{1}{2}(x_{t_{k-1}}-x_{t_{k}}).
\end{equation}\label{eqn:lapSmooth}

Given a time series $\bf{x},$ we will denote by $\bf{\dot{x}}$ and $\bf{\ddot{x}}$ its first and second time derivatives. Our choice is motivated by its compactness and is commonly used in applications involving time derivatives. All derivatives computed are via the usual finite difference numerical approximation.

\subsection{Related Work}

We briefly review the most relevant works to us and otherwise defer the reader to excellent review articles already published: see \cite{aghabozorgi2015time, bagnall2017great, fawaz2019deep} for general overviews. Also, please see \cite{mazimpaka2016trajectory, bian2019trajectory} for trajectory methods.

Analogues of some the specific features we use in our method have been studied within the time series classification literature. In particular, a number of single-model classifiers of interest depend on the extraction of summary features of the behavior of a time series on a given subinterval, sometimes called interval features \cite{rodriguez2004support, deng2013time}. One particular type of interval features of interest, shapelets, are subsequences of given time series that are chosen based on their perceived importance in determining the label of a given time series  \cite{ye2009time,hills2014classification}. Though interval features are natural to consider, they lead to a difficult question of determining which subintervals are most representative of a given class. This is a computationally expensive but necessary question in order to deal with both computational and memory limitations. We avoid these issues by using features that are determined locally around a given time sample, which neither requires a subinterval search nor an extensive amount of memory.

Unsurprisingly, we are not the first to propose a general summary representation for time series. Of recent note is \cite{lubba2019catch22}, in which the authors showed that accurate time series classifiers could be constructed from 22 features that were learned a pool of 4791 possibilities. Our work is distinct in multiple ways: our proposed representation is of fairly higher dimension, does not need to be learned, and is derived only from differential geometric quantities. Perhaps the closest papers to ours in methodology stem from works focused on trajectory classification \cite{zhang2009learning, etemad2018predicting}. Specifically, in~\cite{etemad2018predicting}, the authors extract a number of features from each time point sampled from a given trajectory. These features are then used to compile distributions of each feature that was observed over time. Next, these distributions are summarized by a number of statistics, many of which we employ. The crucial difference in our work is the simultaneous increase the mathematical rigor of their treatment of bearing (the direction in which a trajectory is moving), which is incompatible with quantities of interest from differential geometry. Moreover, our approach is more general and can be applied to time series for which the analogous information is not immediately given. Finally, we propose the use of additional statistical quantities of interest which have nontrivial roles in establishing the accuracy of a classifier based on our presented experiments.

\section{Representing Time Series Through Geometry}\label{sec:pipeline}
\subsection{Review of Differential Geometry for Curves}

As our representation fundamentally relies of differential geometry (and thus, the existence of derivatives), we make the following crucial assumption: \emph{every time series ${\bf{x}}$ of interest is a (sufficiently dense) approximation of a twice continuously differentiable ($C^{2}$) function of time}. By sufficiently dense, we require that numerical first and second derivatives are feasible to compute; both derivatives are explicitly used. This assumption is not restrictive as it is well-known that differentiable functions can well-approximate continuous functions (see, e.g. \cite{stein2009real}). Note that it is possible to transform virtually every time series of interest into one of the assumed form; a combination of linear interpolation and Laplacian smoothing as introduced in Section 2 will suffice, and will be employed later. Note that any noise present within the initial time series may still be present after this transformation, albeit a smoothed, diminished version of it.

Our assumption allows us to consider $d$-dimensional time series ${\bf{x}}$ as a sample of a $C^{2}$ curve in $d+1$-dimensional space: the coordinates at time $t$ of such a curve are $(t,{\bf{x_{t}}}).$ Since we assume the curve is twice differentiable, it is natural to consider its derivatives. Perhaps the most well-known quantities of interest are the velocity:
$${\bf{v_{x}}} := (1,{\bf{\dot{x}}}),$$ 
and its speed $s_{\bf{x}} := ||\bf{v_{x}}||,$ where $||\cdot||$ is the Euclidean norm. Often the first coordinate of $\bf{v_{x}}$ is omitted, reducing to the usual notion of velocity for $d=1.$ Velocity is a natural quantity of interest, as it is well-known that the velocity of a curve along with the value of a curve at any given time is sufficient to uniquely determine the curve; this is a consequence of the fundamental theorem of calculus. It is thus natural to consider $\bf{\dot{v}_{x}},$ more commonly known as \emph{acceleration}. There is another feature of immediate interest: {\emph{curvature}}. 

\begin{definition}The {\emph{curvature vector}} of a $d$-dimensional time series ${\bf{x}}$ is given by $\kappa({\bf{x}}) := {\bf{\dot{v}_{n}}},$ where
\begin{equation}
{\bf{v_{n}}} := \frac{{\bf{v}}}{|| {\bf{v}} ||},
\end{equation}\label{eqn:unitTan}

\end{definition}\label{def:curvature}

As both the name and definition imply, curvature attempts to quantify the change of the direction in which a curve is moving independent of the magnitude of the velocity. The curvature vector is of utmost importance in the study of curves in differential geometry: it is a key component of a {\emph{Darboux frame}}, which is known to uniquely determine a smooth curve up to Euclidean motion \cite{spivak1970comprehensive}. The full power of Darboux frames are well out of scope of this work, though are of great importance and interest in studying the classification of time series on non-Euclidean spaces, which we leave for future work. It is important to note, however that planar curves (e.g. univariate time series) are fully determined up to rigid motion by their curvature vector \cite{do2016differential}. 

For this work, we limit ourselves to $|| \kappa({\bf{x}}) ||,$ known simply as {\emph{curvature}}. Our choice is again motivated by intuition: it is the sharpness of a curve rather than a specific direction that matters more. For univariate time series, one can show

\begin{equation}\label{eqn:curvMag}
|| \kappa({\bf{x}}) || = \frac{|{\bf{\ddot{x}}}|}{||1+{\bf{\dot{x}}}^{2}||^{3}}.
\end{equation}

\noindent Note that this expression gives more intuition as to the meaning of curvature; in the univariate setting, per Equation~\eqref{eqn:curvMag}, it can be thought of as a normalized measure of acceleration, where higher curvature indicates that the curve is bending more. One can also easily define \emph{signed curvature}, where the norm in the numerator of Equation~\eqref{eqn:curvMag} is omitted.
\begin{figure*}[h]
\begin{center}
\includegraphics[scale=.45]{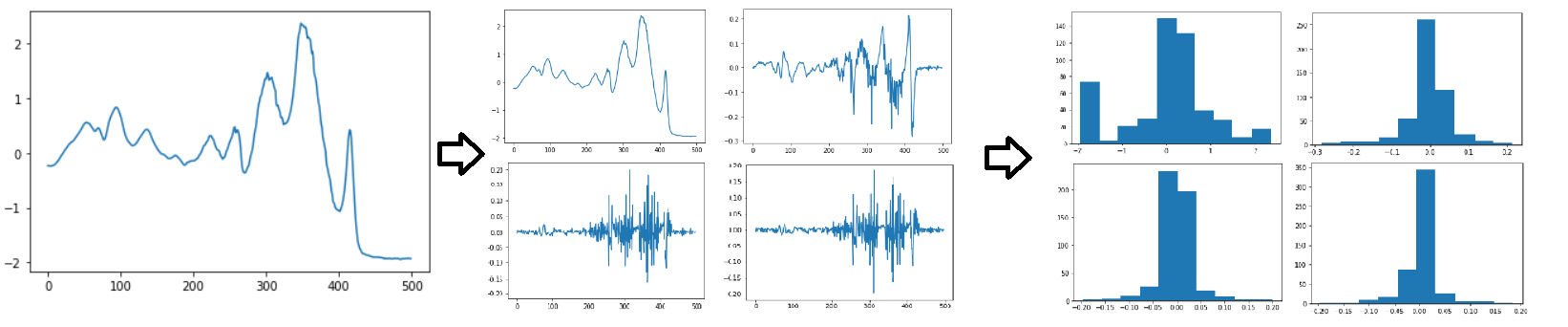}
\end{center}
\caption{Illustration of the GeoDist pipeline. From a given time series, simple derivative-based quantities are extracted, and statistics of their distribution of values are given.}\label{fig:geodist}
\end{figure*}
\subsection{Defining and Extracting GeoStat Representation}

The previous subsection detailed a number of quantities that are mathematically well-understood, are provably useful in determining a (smooth) time series, and are furthermore easy to compute in practice via standard finite difference approximations of derivatives. We wish to use all of these in a formulating the GeoStat (Geometric Statistics). Note that all of the quantities mentioned are functions of time. We proceed by ignoring that they are functions of time and merely consider the \emph{empirical distribution} of each quantity mentioned. From the distributions of each quantity, we extract a sufficient number of statistics from each distribution such that if two sets of said statistics from two different time series are the same, then the distribution of their quantities are approximately equal. 

To be more precise, for a given time series ${\bf{x}},$ we follow the below pipeline for extracting GeoStat representations. The basic idea is illustrated in Figure~\ref{fig:geodist}.

\begin{enumerate}
\item{Upsample ${\bf{x}}$ if necessary using linear interpolation to reach the desired resolution of points}
\item{Apply the desired number of iterations of Laplacian smoothing in Equation~\eqref{eqn:lapSmooth}, compute ${\bf{\dot{x}}}$, and apply the same number of iterations of Laplacian smoothing to ${\bf{\dot{x}}}.$}
\item{Compute ${\bf{\ddot{x}}}, ||{\bf{\ddot{x}}}||,$ and $|| \kappa({\bf{x}}) ||,$ and smooth.}
\item{Extract empirical distributions of relevant features:
\begin{itemize}
\item{For univariate time series: ${\bf{x}}, {\bf{\dot{x}}}, {\bf{\ddot{x}}},$ the curvature, and the signed curvature.}
\item{For multivariate time series: ${\bf{x}}, s_{\bf{x}},$ and $\dot{s}_{\bf{x}}$}
\end{itemize}
and other potentially relevant problem-dependent information, where each sample in the empirical distributions is the value of the corresponding feature at each time.}

\end{enumerate}

For real-valued distributions, we use the following statistics: range, mean, standard deviation, skew, kurtosis, and quantiles. This differs from those used in \cite{etemad2018predicting} in a number of ways. Here, we extract skew and kurtosis in order to capture different notions of spread that are not well captured by the other features. In addition, we also use a more extensive set of quantiles, notably more quantiles towards the tails of each distribution so as to better capture extremal behavior.

For multivariate time series, position is not properly real-valued.  Because of this, the resulting statistics that we can reasonably take are limited; quantiles, for example, do not make sense as there is no natural order to Euclidean space. We instead derive statistics based off of the following quantity:
\begin{equation}
\sigma^{2}(p) = \sum_{i=1}^{n} d^{2}(p_{i},p),
\end{equation}\label{eqn:Frech}
\noindent where the $p_{i}$ are points on the sphere and $d^{2}( \cdot, \cdot )$ is the squared distance between two points on the sphere. The minimum of this function is known as the {\emph{Frechet variance}}, and any $p$ that attains this minimum is known as a {\emph{Frechet mean}}. For Euclidean space, this is the usual mean; this quantity is, however, useful for spherical-valued and other non-Euclidean valued time series. Note that though the Frechet mean is in general not unique, it is if the $p_{i}$ in the summation are within a small enough neighborhood on the sphere \cite{karcher1977riemannian}. In practice, it is easy to compute both quantities by brute force, and we did not run into any issues of nonuniqueness for the data considered. Note that the Frechet variance is always meaningful, even if the Frechet mean is not unique.

Features in GeoStat representations have some degree of sensitivity to various changes in a time series, in the sense that if a time series $\bf{x}$ is transformed into a time series $\bf{F(x)},$ then the features in time series $\bf{F(x)}$ differ from those in $\bf{x},$ albeit generally in a structured way. The resulting change of the quantities mentioned above to various transformations (e.g. offsets, scaling, drift, phase shifts, trend changes) can be analytically determined via derivative properties. Noise will ultimately result in fluctuations in all quantities depending on the structure of the smoothed noise. Per the pipeline, time series with missing values or discontinuities will have those resulting portions interpolated, incorporating these issues into the representation. Time series of unequal duration do not have this explicitly incorporated into their representation, though we encourage the reader to see both the next subsection as well as our discussion on the GFW vessel trajectory dataset \cite{GFWData,GFWWebsite} for ways to address this.
\subsection{Representation by features on windows}

GeoStat representations attempt to give a global summary of a given time series, with no explicit incorporation of notion of locality. This is perhaps a downside, as it has been observed in \cite{fawaz2019deep} that modern deep learning classifiers may learn local features in order to make accurate predictions. Nevertheless, it is possible to easily adapt our proposed representation to incorporate local information via \emph{windowed} GeoStat representaitons: compute all geometric quantities as above, divide the time series into windows, compute the statistical features on each window, and represent the entire time series as a concatenation of all of the features. This also allows for one to address potential unequal durations of time series: once can artificially continue shorter time series (by appending constant values), and then use a multi-window representation. One must take care in using such a representation, as the number of resulting features increases linearly in $k.$ Though how to choose both the number and location of windows used are immediate questions of interest, we do not explicitly pursue this and leave both questions for future work. That being said, we do show that windowed representations using the aforementioned features can result in substantial improvements in classification.
\section{Evaluation and Comparison}\label{sec:comparison}
\subsection{Comparison on the Univariate UCR 2018 Archive}

To illustrate the potential of GeoStat representations, we compare the results of KNN and SVM classifiers trained on these representation with recently proposed deep learning classifiers analyzed in \cite{fawaz2019deep, fawaz2019git}. Our benchmark of interest is the UCR 2018 Time Series Classification Repository \cite{dau2019ucr}, a collection of 128 time series datasets from a variety of applications used to evaluate performance, with each dataset having its own fixed training and test sets. We follow \cite{fawaz2019deep, fawaz2019git} and measure the performance of each classifier by the mean accuracy averaged over five experiments. Note that \cite{fawaz2019deep, fawaz2019git} also reported the mean accuracy averaged over ten experiments for the UCR/UEA Repository, which is itself a subset of the UCR 2018 Repository; for completion, we report the mean accuracy over ten experiments on this smaller collection in the Appendix.

\subsubsection{Experimental setup}
We consider every dataset with the UCR 2018 Repository outside of Crop, for which we ran into memory issues during training. For the small number of datasets in which the individual time series were of unequal length, all time series were upsampled to have the same duration; those with smaller duration were appended with an appropriate number of zeros so as to ensure equal length. Each time series was linearly interpolated, if necessary, to have a minimum of 500 equally spaced samples in time. For each dataset, we construct twelve different feature representations based on the number of iterations of Laplacian smoothing (see Equation~\eqref{eqn:lapSmooth}) and number of windows employed (see the end of Section 3). We specifically consider 0, 1, and 2 iterations of Laplacian smoothing; we also use 1, 2, 4, and 6 windows created by dividing the whole time interval into the corresponding number of consecutive windows of equal length. These representations are then $z$-normalized, and used to train $k$-Nearest Neighbors (KNN) and Support Vector Machine (SVM) classifiers via 10-fold cross validation, with hyperparameters and other relevant quantities given in the Appendix. 

Feature extraction was performed in Python3 using SciPy on a Windows 10 Laptop with a 2.2GHz i7-8750H CPU with 16 GB of RAM. Classification results for KNN and SVM models were obtained using Python3 and SciPy packages on an Ubuntu cluster with four 2.1 GHz Xeon Gold 6152 CPUs and 360 GB of RAM \cite{virtanen2020scipy}. We limit our comparison to other single-model classifiers, namely those in \cite{fawaz2019deep, fawaz2019git} as well as the 1NN dynamic time warping benchmark. We specifically leave out both Time-CNN and t-LeNet due to their poor performance relative to other deep learning models tested, resulting in a total of seven deep learning methods.

\subsubsection{Performance}

\begin{table}[t]

\caption{Ranking of 24 classifiers trained on our representation relative to seven deep learning models and 1-NN DTW on 127 datasets in the UCR 2018 Repository.}
\label{tab:relRank}
\centering
\begin{adjustbox}{max width=0.95\textwidth}
\begin{tabular}{llllllllll}
\toprule
Model & 1st  & 2nd  & 3rd  & 4th  & 5th  & 6th  & 7th  & 8th & 9th  \\
\midrule
KNN\_1W\_0S & 10 &  3 & 22 & 11 & 13 & 12 & 18 & 16 & 22 \\
KNN\_1W\_1S & 12 &  3 & 21 & 15 & 11 & 10 & 15 & 19 & 21 \\
KNN\_1W\_2S & 10 &  3 & 24 & 14 & 14 &  9 & 16 & 14 & 23 \\
KNN\_2W\_0S & 11 & 7 & 25 & 18 & 15 & 13 & 12 & 15 & 11 \\
KNN\_2W\_1S & 14 & 7 & 20 & 19 & 21 & 13 & 13 & 10 & 10 \\
KNN\_2W\_2S & 12 & 7 & 25 & 20 & 21 & 9 & 18 & 9 & 6 \\
KNN\_4W\_0S & 19 & 7 & 26 & 21 & 11 & 11 & 10 & 15 & 7 \\
KNN\_4W\_1S & 16 & 13 & 23 & 21 & 16 & 9 & 13 & 10 & 6 \\
KNN\_4W\_2S & 20 & 10 & 23 & 20 & 12 & 18 & 12 & 8 & 4 \\
KNN\_6W\_0S & 13 & 14 & 24 & 23 & 17 & 10 & 11 & 9 & 6 \\
KNN\_6W\_1S & 17 & 10 & 24 & 22 & 17 & 14 & 14 & 5 & 4 \\
KNN\_6W\_2S & 19 & 10 & 22 & 31 & 11 & 13 & 9 & 7 & 5 \\
\midrule
SVM\_1W\_0S & 16 &  6 & 27 & 17 & 17 & 11 &  7 & 11 & 15 \\
SVM\_1W\_1S & 15 & 10 & 30 & 17 & 11 & 12 & 10 & 10 & 12 \\
SVM\_1W\_2S & 18 & 11 & 26 & 17 & 11 & 12 & 11 & 13 &  8 \\
SVM\_2W\_0S & 15 & 12 & 27 & 29 & 16 & 6 & 10 & 9 & 3 \\
SVM\_2W\_1S & 22 & 12 & 31 & 24 & 9 & 7 & 12 & 9 & 1 \\
SVM\_2W\_2S & 20 & 14 & 35 & 18 & 13 & 11 & 7 & 6 & 3 \\
SVM\_4W\_0S & 21 & 15 & 33 & 20 & 14 & 12 & 5 & 5 & 2 \\
SVM\_4W\_1S & 23 & 17 & 31 & 27 & 8 & 8 & 7 & 4 & 2 \\
SVM\_4W\_2S & 28 & 15 & 32 & 24 & 7 & 9 & 7 & 4 & 1 \\
SVM\_6W\_0S & 23 & 10 & 28 & 23 & 16 & 14 & 8 & 4 & 1 \\
SVM\_6W\_1S & 27 & 13 & 28 & 25 & 14 & 8 & 8 & 3 & 1 \\
SVM\_6W\_2S & 27 & 13 & 36 & 20 & 12 & 10 & 6 & 3 & 0 \\

\midrule
Best KNN & 37 & 12 & 28 & 25 & 8 & 7 & 5 & 3 & 2 \\
Best SVM & 48 & 19 & 34 & 16 & 3 & 4 & 3 & 0 & 0 \\
\midrule
Best Model & 55 & 19 & 28 & 17 & 3 & 3 & 2 & 0 & 0 \\
\bottomrule
\end{tabular}
\end{adjustbox}
\end{table}

Because of the large number of datasets present in this repository, we defer most results to the Appendix and the Supplementary Material. In the main text, we present highlights that reinforce our claim that our representation can be used to create meaningful classifiers.

Table~\ref{tab:relRank} summarizes our results by ranking their average performance over five iterations relative to seven of the deep learning models studied in \cite{fawaz2019deep} and 1-NN DTW.  All models trained under this representation are listed as XXX\_YW\_ZS, where XXX is the model in question, Y is the number of windows used, and Z is the number of Laplacian smoothing iterations used. We also consider, in a separate category, the ranking of the KNN and SVM models with maximum accuracy.  For the sake of transparency and standardization, we obtained performance numbers for the deep learning models from \cite{fawaz2019git}, and we obtained the numbers for 1-NN DTW from the UCR Time Series Repository website \cite{ucr2018website} (note that the procedure used for DTW is deterministic). As \cite{ucr2018website} lists three separate values for DTW (namely, no warping, a learned warping window, and a fixed warping window), we compare to the maximum of these for each dataset. 

One immediately sees how well simple classifiers trained on our representation can perform. In particular, every SVM, as well as all KNN models trained on 4 and 6 window representations, performs better on average than over half of the other eight models considered on every dataset tested. This is perhaps surprising given both the relatively small training complexity of the models we train on our representation, as well as the wide variety of time series within the UCR 2018 Repository. Perhaps most astonishing is that for all but eight of the datasets considered, at least one of the models trained on our representation achieves better classification accuracy than over half of the other models considered. Of the eight datasets for which this is not true, four of the others (FordA, FordB, DodgerLoopGame, DodgerLoopWeekend) have significant high frequency components, which may indicate a weakness of restricting to only differential geometric components, though we do not investigate this further. It is not immediately as to what causes performance issues for the other datasets (Ham, ItalyPowerDemand, SyntheticControl, and TwoPatterns), though we note that at least one of these models is never the worst or second-worst performing per Table~\ref{tab:relRank}. This leaves little question: the particular representation we employ captures significant information necessary to characterize many types of time series.

\begin{table}[t]

\caption{Ranking of the best models trained on our representation relative to each category in the UCR 2018 Repository.}
\label{tab:catRank}
\centering
\begin{adjustbox}{max width=0.95\textwidth}
\begin{tabular}{llllllllll}
\toprule
Category & 1st  & 2nd  & 3rd  & 4th  & 5th  & 6th  & 7th  & 8th & 9th  \\
\midrule
Power & 1 & 0 & 0 & 0 & 0 & 0 & 0 & 0 & 0 \\
HRM & 1 & 0 & 0 & 0 & 0 & 0 & 0 & 0 & 0 \\
Motion & 7 & 5 & 1 & 3 & 0 & 0 & 1 & 0 & 0 \\
Trajectory & 1 & 0 & 2 & 0 & 0 & 0 & 0 & 0 & 0 \\
Simulated & 4 & 1 & 2 & 0 & 0 & 1 & 0 & 0 & 0 \\
Traffic & 0 & 1 & 1 & 0 & 0 & 0 & 0 & 0 & 0 \\
Hemodynamics & 1 & 1 & 1 & 0 & 0 & 0 & 0 & 0 & 0 \\
Spectrum & 3 & 0 & 1 & 0 & 0 & 0 & 0 & 0 & 0 \\
EPG & 2 & 0 & 0 & 0 & 0 & 0 & 0 & 0 & 0 \\
Spectro & 5 & 2 & 1 & 0 & 0 & 0 & 0 & 0 & 0 \\
EOG & 0 & 1 & 0 & 1 & 0 & 0 & 0 & 0 & 0 \\
ECG & 2 & 1 & 2 & 0 & 1 & 0 & 0 & 0 & 0 \\
Sensor & 8 & 2 & 10 & 7 & 2 & 1 & 0 & 0 & 0 \\
Device & 5 & 1 & 1 & 0 & 0 & 1 & 0 & 0 & 0 \\
Image & 15 & 4 & 6 & 6 & 0 & 0 & 1 & 0 & 0 \\
\bottomrule
\end{tabular}
\end{adjustbox}
\end{table}

It is natural to ask if the rankings in Table~\ref{tab:relRank} are biased towards a particular type of data. We list this information in Table~\ref{tab:catRank}, where the categories are those listed in the UCR 2018 Repository. The only category for which the best GeoStat model is neither first nor second for the majority of the datasets is Sensor, giving credence to a potential weakness of our representation being significant high-frequency components. Other categories have one of the models trained on GeoStat representations within the top 2 of those considered.

We conclude this section with two remarks. First, we note per the Appendix that GeoStat representations allows for perfect classification for the given train/test splits on a number of datasets (e.g. Beef, BeetleFly, and BirdChicken) that have proven difficult for the other methods in our comparison. Second, given that we are proposing GeoStat representation in part as a reasonable DTW-free benchmark, we note that at least one of the models trained on our representation has average accuracy equal or better than that of the DTW accuracy we list on 106 of the 127 datasets considered. Though not perfect, given the relative simplicity of our methods, this does suggest that GeoStat representations can be used as such a benchmark provided sufficient hyperparameter tuning.
\subsection{Multivariate Case Study: Vessel Classification}

\begin{table}[]
\caption{Statistics of 10-fold nested cross validation scores for KNN and SVM with respect to the given features.}
\label{tab:transshipAcc}
\centering
\begin{adjustbox}{max width=0.95\textwidth}
\begin{tabular}{lllll}
\toprule
Model & Min      & Max         & Mean        & Std      \\
\midrule
KNN   & 0.6549 & 0.6883 & 0.6701 & 0.0079 \\
SVM   & 0.6810 & 0.7028 & 0.6921 & 0.0061 \\
\bottomrule
\end{tabular}
\end{adjustbox}
\end{table}

As previously mentioned in Section 1, we will evaluate the performance of our method in the multivariate setting on a challenging dataset from Global Fishing Watch \cite{GFWData,GFWWebsite}. This dataset consists of 1258 vessel trajectories spanning the entire globe (a sphere) over a period of four years. Figure~\ref{fig:spatialRange} shows the spatial range of one hundred of these trajectories (with some GPS errors) over this time period. Of these, 1107 are labeled with one of twelve different vessel classes. The goal is accurate classification. The current state of the art on this problem given in \cite{kroodsma2018tracking} is a large CNN that was trained and tested on a much larger superset of the data freely available; we have approximately 2\% of what was used. Their model achieves a high classification accuracy of approximately 95\% on a fixed training set of 52,964 and a test set of 22,172. We are interested in seeing if the features employed combined with SVM and KNN models can yield good performance despite a lack of data. Note that this dataset requires a fair amount of preprocessing before we use it; details are given in the Appendix. In particular, irregular sampling rates are relevant to vessel behavior \cite{kroodsma2018tracking}.

Given the geographic scope of the data, the dataset that we have to work with is itself relatively small. We ideally need access to as much data as possible, but formally holding out a fixed dataset would perhaps highly bias our result. We instead use 10-fold {\emph{nested cross-validation}}, which allows us to generate random train-validation-test splits. To be more precise, we will first split the dataset into 10 parts of equal size. One of these parts will act as a holdout set. The other nine parts will be combined and then split into another 10 pieces, which we will run cross-validation on to tune hyperparameters (the same as in the previous section). Based on these chosen hyperparameters, the final loss is evaluated on the holdout set from the first split. This is repeated so that each of the folds is used as a holdout set; the final loss is the average of the losses for each holdout set. 

We performed thirty iterations of this 10-fold nested cross validation procedure. Statistics of these iterations are given in Table~\ref{tab:transshipAcc}. Both models perform surprisingly well in this scenario, with KNN and SVM achieving 67\% and 69\% accuracy on average, respectively, whereas approximately 50-times as much data with a deep learning model achieves 95\% accuracy for a fixed train/test split. We give a confusion matrix in the Appendix for a fixed nested cross-validation iteration. In particular, note that a large amount of inaccuracy comes from confusing certain classes of vessels, such as reefers and cargo/tankers, which have similar movement patterns and are potentially mistakable by just trajectory information alone.
\section{Empirical Studies}\label{sec:parameter_eval}
We discuss particular choices made in the course of constructing the presented GeoStat representations. Though some discussions will be based off of material presented in the main text, full details of many will be deferred to the Appendix.

\subsection{Amount of Laplacian smoothings} Based on Table~\ref{tab:relRank}, we see a general pattern of improved performance with additional iterations of smoothing. This is not true of every model: KNN\_1W\_1S generally performs better than KNN\_1W\_2S. The general pattern, however, is perhaps not surprising; the Laplacian smoothing we employed effectively acts as a denoising algorithm.

\subsection{Amount of Windows Used} A perhaps notable trend is the role of using more windows. By representing a time series by multiple windows, performance can significantly increase. This is most evident by comparing the 1-window SVM models to the 6-window SVM models. We again remark that our choice of windows was not algorithmic, and leave the possibility that one can learn a good choice of windows as future work.

\subsection{Effect of specific windows}

As discussed in \cite{fawaz2019deep}, the success of deep learning on time series classification might be attributed to the ability to learn localized features; the performance of KNN and SVM with multiple windows suggests the same. To test this, we trained both KNN and SVM classifiers on the statistics from each of the individual windows extracted on the 6-window dataset with one smoothing iteration. Though restricting to individual windows does not appear to improve performance, near-perfect or perfect classification results can be achieved using only information from specific windows (e.g. Beef, BeetleFly, BirdChicken, and StarlightCurves). This suggests that future improvements on classification may rely on the ability for models to find local areas of interest, at least in certain cases. Other results show (e.g. on the Adiac and Cricket), however, that this is not true in general.

\subsection{Parameter ablation studies} We conducted parameter ablation studies on each of the datasets studied. Specifically, we tested the effect of removing specific geometric features (e.g. position, curvature) and specific statistics on classifier performance. All tables are given in the Appendix, with the datasets from the UCR 2018 Repository separated into the categories given in Table~\ref{tab:catRank} for brevity. Given both the breadth and the unequal amount of datasets per category, it is difficult to jump to conclusions; both the HRM and Power categories, for example, only have one dataset. The importance of both the position and velocity information is clear throughout, resulting in significant accuracy decreases when omitted. Similarly, extremal quantiles also appear to be beneficial for classifying time series in general. Perhaps the most contentious quantities involve second derivative information, being detrimental for some categories and important for others, noting again that some categories have very few datasets. Given this, we can only conclude the importance of position, velocity, and extremal quantile information in general, suggesting that the importance of other parameters be evaluated on a case by case basis.

\section{Conclusion}\label{sec:conclusion}
We proposed GeoStat representations for time series, summarizing time series by summary statistics of distributions of geometric values. We showed that simple classifiers trained on these representations can be extremely competitive with modern single-model classifiers based on deep learning, as well as the gold standard 1NN-DTW benchmark, achieving state of the art performance on a number of real datasets. We also showed that such methods achieve good performance on the difficult, limited multivariate GFW dataset relative to that of the state of the art despite the small amount of data.

There are many directions for future work. As noted in Section 4, our representation may not be sufficient for characterizing time series with significant high frequency components. It is not clear if purely geometric quantities can address this, but doing so, preferably with well-understood mathematical quantities, is an immediate direction for future work. The particular choice of representing statistics lends itself to interesting questions. It is interesting to ask whether a more compact, or more mathematically natural representation of said distributions would have similar or better performance. Finally, as we discussed earlier, we did little work in investigating window choices. Though some of our results on restriction to particular windows were enlightening, further studies should be done to determine particular regions of interest for classification.

\bibliographystyle{abbrv}
\bibliography{biblio}
\newpage
\section{Appendix}\label{sec:appendix}
\subsection{Supplementary Material structure}
The Supplementary Material is divided into two folders, one for the univariate experiments and one for the multivariate experiments. The results folder contains all data used for every result presented in the main paper and the Appendix, along with additional information for each set of experiments performed. Every model and window combination contains, for each dataset, the average accuracy over all runs of the same experiment, as well as the minimum/maximum accuracies and the standard deviation. These do not include the results from \cite{fawaz2019deep} or \cite{ucr2018website}, which themselves are available on \cite{fawaz2019git} and \cite{ucr2018website} respectively. We do, however, include the script that was used to extract the results from the raw results posted on \cite{fawaz2019git}; the results from \cite{ucr2018website} were downloaded from the website and manually processed.

We include the processed data from the multivariate experiment, but due to size constraints, do not include any of the raw data nor any of the processed UCR data. The raw data can be obtained from \cite{GFWData} and \cite{ucr2018website}, and we include all extraction scripts (with listed order for running) used in obtaining them. Those interested in doing so should only need to specify paths. However, it is important to note that both raw datasets (and the processed UCR dataset) are quite large. This may render personal verification infeasible without sufficient resources, though please note that none of the results were derived with GPUs.

\subsection{Hyperparameters}
The following hyperparameters were used for both univariate and multivariate tests. All variable names follow the same convention as in \cite{virtanen2020scipy}
\begin{itemize}
    \item{KNN \begin{itemize}
        \item{n\textunderscore neighbors: 1, 2, 4, 6, 8, 10}
        \item{weights: 'uniform','distance'}
        \item{p: 1,2}
    \end{itemize}}
    \item{SVM \begin{itemize}
        \item{C: 0.1, 1, 10}
        \item{kernel: 'linear','rbf','poly'}
        \item{degree: 2}
    \end{itemize}}
\end{itemize}
\subsection{Univariate comparison results}
\subsubsection{Quantiles}
We employ the following quantiles: 0.001, 0.01, 0.1, 0.2, 0.3, 0.4, 0.5, 0.6, 0.7, 0.8, 0.9, 0.99, 0.999.

\subsubsection{UCR 2018 Results}
\begin{table*}[]
\caption{Tables showing accuracy statistics for the UCR 2018 repository with five iterations of cross validation. The results listed here are for SVM (left) and KNN (right) trained on 4-window GeoStat representations with two iterations of Laplacian smoothing.}
\label{tab:UCR2018Full}

\begin{center}
\resizebox*{0.65\textwidth}{0.95\textheight}{
\begin{tabular}{@{}lllll@{}}
\toprule
Dataset (SVM)                          & Min    & Max    & Mean   & Std. Dev  \\
\midrule
ACSF1                          & 0.6200 & 0.6500 & 0.6260 & 0.0120 \\
Adiac                          & 0.7775 & 0.7826 & 0.7785 & 0.0020 \\
AllGestureWiimoteX             & 0.5471 & 0.5471 & 0.5471 & 0.0000 \\
AllGestureWiimoteY             & 0.5900 & 0.5900 & 0.5900 & 0.0000 \\
AllGestureWiimoteZ             & 0.5500 & 0.5500 & 0.5500 & 0.0000 \\
ArrowHead                      & 0.7371 & 0.7371 & 0.7371 & 0.0000 \\
BME                            & 0.9867 & 0.9867 & 0.9867 & 0.0000 \\
Beef                           & 0.7667 & 0.7667 & 0.7667 & 0.0000 \\
BeetleFly                      & 0.5000 & 0.9000 & 0.7100 & 0.1319 \\
BirdChicken                    & 0.9000 & 0.9000 & 0.9000 & 0.0000 \\
CBF                            & 0.9756 & 0.9756 & 0.9756 & 0.0000 \\
Car                            & 0.8333 & 0.8833 & 0.8533 & 0.0245 \\
Chinatown                      & 0.7522 & 0.7522 & 0.7522 & 0.0000 \\
ChlorineConcentration          & 0.6901 & 0.6901 & 0.6901 & 0.0000 \\
CinCECGTorso                   & 0.9246 & 0.9246 & 0.9246 & 0.0000 \\
Coffee                         & 1.0000 & 1.0000 & 1.0000 & 0.0000 \\
Computers                      & 0.6000 & 0.6040 & 0.6032 & 0.0016 \\
CricketX                       & 0.6487 & 0.6487 & 0.6487 & 0.0000 \\
CricketY                       & 0.7000 & 0.7000 & 0.7000 & 0.0000 \\
CricketZ                       & 0.6359 & 0.6846 & 0.6749 & 0.0195 \\
DiatomSizeReduction            & 0.9608 & 0.9608 & 0.9608 & 0.0000 \\
DistalPhalanxOutlineAgeGroup   & 0.7338 & 0.7338 & 0.7338 & 0.0000 \\
DistalPhalanxOutlineCorrect    & 0.7681 & 0.7717 & 0.7703 & 0.0018 \\
DistalPhalanxTW                & 0.6691 & 0.6691 & 0.6691 & 0.0000 \\
DodgerLoopDay                  & 0.3500 & 0.4125 & 0.3625 & 0.0250 \\
DodgerLoopGame                 & 0.7391 & 0.7536 & 0.7507 & 0.0058 \\
DodgerLoopWeekend              & 0.9710 & 0.9710 & 0.9710 & 0.0000 \\
ECG200                         & 0.8100 & 0.8400 & 0.8160 & 0.0120 \\
ECG5000                        & 0.9353 & 0.9362 & 0.9357 & 0.0004 \\
ECGFiveDays                    & 0.9501 & 0.9501 & 0.9501 & 0.0000 \\
EOGHorizontalSignal            & 0.4890 & 0.5028 & 0.4917 & 0.0055 \\
EOGVerticalSignal              & 0.4420 & 0.4448 & 0.4425 & 0.0011 \\
Earthquakes                    & 0.7338 & 0.7482 & 0.7424 & 0.0070 \\
ElectricDevices                & 0.7041 & 0.7041 & 0.7041 & 0.0000 \\
EthanolLevel                   & 0.7580 & 0.7660 & 0.7644 & 0.0032 \\
FaceAll                        & 0.8018 & 0.8018 & 0.8018 & 0.0000 \\
FaceFour                       & 0.7841 & 0.7841 & 0.7841 & 0.0000 \\
FacesUCR                       & 0.8610 & 0.8610 & 0.8610 & 0.0000 \\
FiftyWords                     & 0.7604 & 0.7670 & 0.7618 & 0.0026 \\
Fish                           & 0.9257 & 0.9257 & 0.9257 & 0.0000 \\
FordA                          & 0.8242 & 0.8242 & 0.8242 & 0.0000 \\
FordB                          & 0.6617 & 0.6852 & 0.6711 & 0.0115 \\
FreezerRegularTrain            & 0.9947 & 0.9947 & 0.9947 & 0.0000 \\
FreezerSmallTrain              & 0.9744 & 0.9744 & 0.9744 & 0.0000 \\
Fungi                          & 0.8602 & 0.8602 & 0.8602 & 0.0000 \\
GestureMidAirD1                & 0.6615 & 0.6615 & 0.6615 & 0.0000 \\
GestureMidAirD2                & 0.4923 & 0.5538 & 0.5046 & 0.0246 \\
GestureMidAirD3                & 0.3231 & 0.3231 & 0.3231 & 0.0000 \\
GesturePebbleZ1                & 0.8430 & 0.8837 & 0.8756 & 0.0163 \\
GesturePebbleZ2                & 0.7975 & 0.7975 & 0.7975 & 0.0000 \\
GunPoint                       & 1.0000 & 1.0000 & 1.0000 & 0.0000 \\
GunPointAgeSpan                & 0.9842 & 0.9842 & 0.9842 & 0.0000 \\
GunPointMaleVersusFemale       & 0.9842 & 0.9937 & 0.9892 & 0.0038 \\
GunPointOldVersusYoung         & 1.0000 & 1.0000 & 1.0000 & 0.0000 \\
Ham                            & 0.6571 & 0.6762 & 0.6648 & 0.0093 \\
HandOutlines                   & 0.9324 & 0.9324 & 0.9324 & 0.0000 \\
Haptics                        & 0.4968 & 0.5195 & 0.5149 & 0.0091 \\
Herring                        & 0.6250 & 0.6406 & 0.6313 & 0.0077 \\
HouseTwenty                    & 0.9412 & 0.9664 & 0.9563 & 0.0124 \\
InlineSkate                    & 0.4891 & 0.5291 & 0.5211 & 0.0160 \\
InsectEPGRegularTrain          & 1.0000 & 1.0000 & 1.0000 & 0.0000 \\
InsectEPGSmallTrain            & 0.9719 & 0.9719 & 0.9719 & 0.0000 \\
InsectWingbeatSound            & 0.5354 & 0.5753 & 0.5515 & 0.0194 \\
ItalyPowerDemand               & 0.9378 & 0.9466 & 0.9396 & 0.0035 \\
LargeKitchenAppliances         & 0.7200 & 0.7200 & 0.7200 & 0.0000 \\
Lightning2                     & 0.7377 & 0.7705 & 0.7443 & 0.0131 \\
Lightning7                     & 0.6438 & 0.6712 & 0.6603 & 0.0134 \\
Mallat                         & 0.9390 & 0.9390 & 0.9390 & 0.0000 \\
Meat                           & 0.9500 & 0.9500 & 0.9500 & 0.0000 \\
MedicalImages                  & 0.7092 & 0.7382 & 0.7324 & 0.0116 \\
MelbournePedestrian            & 0.8963 & 0.9008 & 0.8991 & 0.0021 \\
MiddlePhalanxOutlineAgeGroup   & 0.5260 & 0.6234 & 0.6039 & 0.0390 \\
MiddlePhalanxOutlineCorrect    & 0.8351 & 0.8385 & 0.8371 & 0.0017 \\
MiddlePhalanxTW                & 0.5974 & 0.5974 & 0.5974 & 0.0000 \\
MixedShapesRegularTrain        & 0.9559 & 0.9654 & 0.9578 & 0.0038 \\
MixedShapesSmallTrain          & 0.9320 & 0.9320 & 0.9320 & 0.0000 \\
MoteStrain                     & 0.8546 & 0.8546 & 0.8546 & 0.0000 \\
NonInvasiveFetalECGThorax1     & 0.9226 & 0.9252 & 0.9232 & 0.0010 \\
NonInvasiveFetalECGThorax2     & 0.9257 & 0.9262 & 0.9259 & 0.0002 \\
OSULeaf                        & 0.8182 & 0.8182 & 0.8182 & 0.0000 \\
OliveOil                       & 0.9000 & 0.9000 & 0.9000 & 0.0000 \\
PLAID                          & 0.6946 & 0.7002 & 0.6968 & 0.0018 \\
PhalangesOutlinesCorrect       & 0.8077 & 0.8077 & 0.8077 & 0.0000 \\
Phoneme                        & 0.2959 & 0.2959 & 0.2959 & 0.0000 \\
PickupGestureWiimoteZ          & 0.6600 & 0.6600 & 0.6600 & 0.0000 \\
PigAirwayPressure              & 0.1587 & 0.1587 & 0.1587 & 0.0000 \\
PigArtPressure                 & 0.8462 & 0.8462 & 0.8462 & 0.0000 \\
PigCVP                         & 0.4423 & 0.4423 & 0.4423 & 0.0000 \\
Plane                          & 1.0000 & 1.0000 & 1.0000 & 0.0000 \\
PowerCons                      & 0.9389 & 0.9389 & 0.9389 & 0.0000 \\
ProximalPhalanxOutlineAgeGroup & 0.8585 & 0.8585 & 0.8585 & 0.0000 \\
ProximalPhalanxOutlineCorrect  & 0.8454 & 0.8625 & 0.8488 & 0.0069 \\
ProximalPhalanxTW              & 0.7707 & 0.8146 & 0.7971 & 0.0215 \\
RefrigerationDevices           & 0.5653 & 0.5653 & 0.5653 & 0.0000 \\
Rock                           & 0.8000 & 0.8000 & 0.8000 & 0.0000 \\
ScreenType                     & 0.4507 & 0.4507 & 0.4507 & 0.0000 \\
SemgHandGenderCh2              & 0.8833 & 0.9000 & 0.8867 & 0.0067 \\
SemgHandMovementCh2            & 0.7178 & 0.7178 & 0.7178 & 0.0000 \\
SemgHandSubjectCh2             & 0.8533 & 0.8533 & 0.8533 & 0.0000 \\
ShakeGestureWiimoteZ           & 0.8800 & 0.8800 & 0.8800 & 0.0000 \\
ShapeletSim                    & 0.6278 & 0.6500 & 0.6411 & 0.0109 \\
ShapesAll                      & 0.8350 & 0.8350 & 0.8350 & 0.0000 \\
SmallKitchenAppliances         & 0.7920 & 0.7920 & 0.7920 & 0.0000 \\
SmoothSubspace                 & 0.9533 & 0.9600 & 0.9560 & 0.0033 \\
SonyAIBORobotSurface1          & 0.8419 & 0.8419 & 0.8419 & 0.0000 \\
SonyAIBORobotSurface2          & 0.8940 & 0.8940 & 0.8940 & 0.0000 \\
StarLightCurves                & 0.9772 & 0.9772 & 0.9772 & 0.0000 \\
Strawberry                     & 0.9541 & 0.9649 & 0.9627 & 0.0043 \\
SwedishLeaf                    & 0.9584 & 0.9584 & 0.9584 & 0.0000 \\
Symbols                        & 0.9618 & 0.9618 & 0.9618 & 0.0000 \\
SyntheticControl               & 0.9600 & 0.9600 & 0.9600 & 0.0000 \\
ToeSegmentation1               & 0.8509 & 0.8509 & 0.8509 & 0.0000 \\
ToeSegmentation2               & 0.7462 & 0.7769 & 0.7523 & 0.0123 \\
Trace                          & 0.9900 & 0.9900 & 0.9900 & 0.0000 \\
TwoLeadECG                     & 0.9860 & 0.9860 & 0.9860 & 0.0000 \\
TwoPatterns                    & 0.9708 & 0.9708 & 0.9708 & 0.0000 \\
UMD                            & 0.9931 & 0.9931 & 0.9931 & 0.0000 \\
UWaveGestureLibraryAll         & 0.9453 & 0.9497 & 0.9489 & 0.0018 \\
UWaveGestureLibraryX           & 0.8082 & 0.8082 & 0.8082 & 0.0000 \\
UWaveGestureLibraryY           & 0.7457 & 0.7457 & 0.7457 & 0.0000 \\
UWaveGestureLibraryZ           & 0.7440 & 0.7440 & 0.7440 & 0.0000 \\
Wafer                          & 0.9927 & 0.9927 & 0.9927 & 0.0000 \\
Wine                           & 0.5741 & 0.8704 & 0.6926 & 0.1452 \\
WordSynonyms                   & 0.6411 & 0.6897 & 0.6605 & 0.0238 \\
Worms                          & 0.7143 & 0.7143 & 0.7143 & 0.0000 \\
WormsTwoClass                  & 0.7532 & 0.8052 & 0.7948 & 0.0208 \\
Yoga                           & 0.8623 & 0.8623 & 0.8623 & 0.0000   \\
\bottomrule
\end{tabular}
\:\:\:\:\:\:\:\:\:
\begin{tabular}{@{}lllll@{}}
\toprule
Dataset (KNN)                          & Min    & Max    & Mean   & Std. Dev \\
\midrule
ACSF1                          & 0.7600 & 0.7600 & 0.7600 & 0.0000 \\
Adiac                          & 0.7084 & 0.7084 & 0.7084 & 0.0000 \\
AllGestureWiimoteX             & 0.5800 & 0.5800 & 0.5800 & 0.0000 \\
AllGestureWiimoteY             & 0.5414 & 0.6171 & 0.6020 & 0.0303 \\
AllGestureWiimoteZ             & 0.5971 & 0.5971 & 0.5971 & 0.0000 \\
ArrowHead                      & 0.7886 & 0.7886 & 0.7886 & 0.0000 \\
BME                            & 0.9600 & 0.9600 & 0.9600 & 0.0000 \\
Beef                           & 0.5667 & 0.5667 & 0.5667 & 0.0000 \\
BeetleFly                      & 0.7000 & 0.8000 & 0.7400 & 0.0490 \\
BirdChicken                    & 0.8000 & 0.8000 & 0.8000 & 0.0000 \\
CBF                            & 0.9656 & 0.9711 & 0.9667 & 0.0022 \\
Car                            & 0.7167 & 0.7833 & 0.7533 & 0.0245 \\
Chinatown                      & 0.9242 & 0.9679 & 0.9592 & 0.0175 \\
ChlorineConcentration          & 0.6693 & 0.6693 & 0.6693 & 0.0000 \\
CinCECGTorso                   & 0.8406 & 0.8406 & 0.8406 & 0.0000 \\
Coffee                         & 0.9643 & 1.0000 & 0.9714 & 0.0143 \\
Computers                      & 0.6080 & 0.6480 & 0.6272 & 0.0139 \\
CricketX                       & 0.6231 & 0.6231 & 0.6231 & 0.0000 \\
CricketY                       & 0.6359 & 0.6359 & 0.6359 & 0.0000 \\
CricketZ                       & 0.6282 & 0.6436 & 0.6333 & 0.0056 \\
DiatomSizeReduction            & 0.9379 & 0.9379 & 0.9379 & 0.0000 \\
DistalPhalanxOutlineAgeGroup   & 0.7122 & 0.7194 & 0.7137 & 0.0029 \\
DistalPhalanxOutlineCorrect    & 0.7500 & 0.7500 & 0.7500 & 0.0000 \\
DistalPhalanxTW                & 0.6331 & 0.6475 & 0.6446 & 0.0058 \\
DodgerLoopDay                  & 0.3625 & 0.4500 & 0.4100 & 0.0278 \\
DodgerLoopGame                 & 0.7464 & 0.7464 & 0.7464 & 0.0000 \\
DodgerLoopWeekend              & 0.9493 & 0.9710 & 0.9609 & 0.0098 \\
ECG200                         & 0.8000 & 0.8000 & 0.8000 & 0.0000 \\
ECG5000                        & 0.9413 & 0.9433 & 0.9421 & 0.0010 \\
ECGFiveDays                    & 0.8955 & 0.8955 & 0.8955 & 0.0000 \\
EOGHorizontalSignal            & 0.3619 & 0.4061 & 0.3796 & 0.0165 \\
EOGVerticalSignal              & 0.3177 & 0.3591 & 0.3508 & 0.0166 \\
Earthquakes                    & 0.7338 & 0.7482 & 0.7424 & 0.0070 \\
ElectricDevices                & 0.6631 & 0.6681 & 0.6659 & 0.0023 \\
EthanolLevel                   & 0.4240 & 0.4240 & 0.4240 & 0.0000 \\
FaceAll                        & 0.7550 & 0.8515 & 0.7936 & 0.0473 \\
FaceFour                       & 0.7727 & 0.8409 & 0.7864 & 0.0273 \\
FacesUCR                       & 0.8059 & 0.8420 & 0.8343 & 0.0143 \\
FiftyWords                     & 0.7187 & 0.7187 & 0.7187 & 0.0000 \\
Fish                           & 0.8514 & 0.8743 & 0.8663 & 0.0078 \\
FordA                          & 0.7174 & 0.7197 & 0.7183 & 0.0011 \\
FordB                          & 0.6827 & 0.6889 & 0.6872 & 0.0024 \\
FreezerRegularTrain            & 0.9544 & 0.9740 & 0.9667 & 0.0068 \\
FreezerSmallTrain              & 0.9288 & 0.9309 & 0.9292 & 0.0008 \\
Fungi                          & 0.9409 & 0.9409 & 0.9409 & 0.0000 \\
GestureMidAirD1                & 0.5923 & 0.6231 & 0.6046 & 0.0151 \\
GestureMidAirD2                & 0.5077 & 0.5077 & 0.5077 & 0.0000 \\
GestureMidAirD3                & 0.2385 & 0.2769 & 0.2523 & 0.0141 \\
GesturePebbleZ1                & 0.8430 & 0.8547 & 0.8512 & 0.0047 \\
GesturePebbleZ2                & 0.6519 & 0.7658 & 0.7114 & 0.0362 \\
GunPoint                       & 0.9933 & 0.9933 & 0.9933 & 0.0000 \\
GunPointAgeSpan                & 1.0000 & 1.0000 & 1.0000 & 0.0000 \\
GunPointMaleVersusFemale       & 0.9968 & 0.9968 & 0.9968 & 0.0000 \\
GunPointOldVersusYoung         & 1.0000 & 1.0000 & 1.0000 & 0.0000 \\
Ham                            & 0.5905 & 0.6000 & 0.5981 & 0.0038 \\
HandOutlines                   & 0.8892 & 0.8919 & 0.8897 & 0.0011 \\
Haptics                        & 0.4513 & 0.4740 & 0.4669 & 0.0083 \\
Herring                        & 0.5938 & 0.7500 & 0.6750 & 0.0636 \\
HouseTwenty                    & 0.9412 & 0.9412 & 0.9412 & 0.0000 \\
InlineSkate                    & 0.5782 & 0.5782 & 0.5782 & 0.0000 \\
InsectEPGRegularTrain          & 0.9920 & 0.9920 & 0.9920 & 0.0000 \\
InsectEPGSmallTrain            & 0.9880 & 0.9880 & 0.9880 & 0.0000 \\
InsectWingbeatSound            & 0.4808 & 0.5005 & 0.4927 & 0.0068 \\
ItalyPowerDemand               & 0.9388 & 0.9388 & 0.9388 & 0.0000 \\
LargeKitchenAppliances         & 0.6640 & 0.6960 & 0.6768 & 0.0157 \\
Lightning2                     & 0.7377 & 0.8197 & 0.8033 & 0.0328 \\
Lightning7                     & 0.6712 & 0.6712 & 0.6712 & 0.0000 \\
Mallat                         & 0.9684 & 0.9684 & 0.9684 & 0.0000 \\
Meat                           & 0.9000 & 0.9500 & 0.9367 & 0.0194 \\
MedicalImages                  & 0.6526 & 0.6974 & 0.6845 & 0.0161 \\
MelbournePedestrian            & 0.9147 & 0.9184 & 0.9162 & 0.0018 \\
MiddlePhalanxOutlineAgeGroup   & 0.5455 & 0.5909 & 0.5610 & 0.0195 \\
MiddlePhalanxOutlineCorrect    & 0.7904 & 0.7904 & 0.7904 & 0.0000 \\
MiddlePhalanxTW                & 0.5390 & 0.5779 & 0.5584 & 0.0123 \\
MixedShapesRegularTrain        & 0.9588 & 0.9588 & 0.9588 & 0.0000 \\
MixedShapesSmallTrain          & 0.9229 & 0.9328 & 0.9308 & 0.0040 \\
MoteStrain                     & 0.8578 & 0.8578 & 0.8578 & 0.0000 \\
NonInvasiveFetalECGThorax1     & 0.8656 & 0.8687 & 0.8663 & 0.0012 \\
NonInvasiveFetalECGThorax2     & 0.8707 & 0.8728 & 0.8711 & 0.0008 \\
OSULeaf                        & 0.6983 & 0.8140 & 0.7744 & 0.0430 \\
OliveOil                       & 0.8333 & 0.9333 & 0.8867 & 0.0452 \\
PLAID                          & 0.7486 & 0.7747 & 0.7695 & 0.0104 \\
PhalangesOutlinesCorrect       & 0.7821 & 0.7890 & 0.7862 & 0.0034 \\
Phoneme                        & 0.2126 & 0.2748 & 0.2563 & 0.0224 \\
PickupGestureWiimoteZ          & 0.5400 & 0.6800 & 0.5840 & 0.0571 \\
PigAirwayPressure              & 0.0962 & 0.1731 & 0.1577 & 0.0308 \\
PigArtPressure                 & 0.9038 & 0.9038 & 0.9038 & 0.0000 \\
PigCVP                         & 0.4471 & 0.4471 & 0.4471 & 0.0000 \\
Plane                          & 1.0000 & 1.0000 & 1.0000 & 0.0000 \\
PowerCons                      & 0.8611 & 0.9056 & 0.8933 & 0.0163 \\
ProximalPhalanxOutlineAgeGroup & 0.8390 & 0.8585 & 0.8507 & 0.0073 \\
ProximalPhalanxOutlineCorrect  & 0.8179 & 0.8213 & 0.8192 & 0.0017 \\
ProximalPhalanxTW              & 0.7902 & 0.8098 & 0.8010 & 0.0089 \\
RefrigerationDevices           & 0.4747 & 0.5173 & 0.4917 & 0.0209 \\
Rock                           & 0.6600 & 0.6600 & 0.6600 & 0.0000 \\
ScreenType                     & 0.4187 & 0.4533 & 0.4384 & 0.0161 \\
SemgHandGenderCh2              & 0.9267 & 0.9483 & 0.9353 & 0.0106 \\
SemgHandMovementCh2            & 0.7978 & 0.8000 & 0.7996 & 0.0009 \\
SemgHandSubjectCh2             & 0.8444 & 0.8600 & 0.8520 & 0.0064 \\
ShakeGestureWiimoteZ           & 0.7400 & 0.7400 & 0.7400 & 0.0000 \\
ShapeletSim                    & 0.5833 & 0.6611 & 0.6322 & 0.0286 \\
ShapesAll                      & 0.8433 & 0.8433 & 0.8433 & 0.0000 \\
SmallKitchenAppliances         & 0.7467 & 0.7467 & 0.7467 & 0.0000 \\
SmoothSubspace                 & 0.9467 & 0.9800 & 0.9547 & 0.0129 \\
SonyAIBORobotSurface1          & 0.8602 & 0.8602 & 0.8602 & 0.0000 \\
SonyAIBORobotSurface2          & 0.7754 & 0.8006 & 0.7943 & 0.0098 \\
StarLightCurves                & 0.9720 & 0.9751 & 0.9735 & 0.0013 \\
Strawberry                     & 0.9486 & 0.9541 & 0.9519 & 0.0026 \\
SwedishLeaf                    & 0.9312 & 0.9312 & 0.9312 & 0.0000 \\
Symbols                        & 0.9397 & 0.9397 & 0.9397 & 0.0000 \\
SyntheticControl               & 0.9233 & 0.9233 & 0.9233 & 0.0000 \\
ToeSegmentation1               & 0.6228 & 0.7325 & 0.7044 & 0.0425 \\
ToeSegmentation2               & 0.7538 & 0.7846 & 0.7785 & 0.0123 \\
Trace                          & 1.0000 & 1.0000 & 1.0000 & 0.0000 \\
TwoLeadECG                     & 0.8060 & 0.9298 & 0.8325 & 0.0488 \\
TwoPatterns                    & 0.7553 & 0.7718 & 0.7675 & 0.0062 \\
UMD                            & 0.8958 & 0.9792 & 0.9556 & 0.0327 \\
UWaveGestureLibraryAll         & 0.9196 & 0.9327 & 0.9222 & 0.0052 \\
UWaveGestureLibraryX           & 0.7889 & 0.8102 & 0.8012 & 0.0100 \\
UWaveGestureLibraryY           & 0.7384 & 0.7471 & 0.7453 & 0.0035 \\
UWaveGestureLibraryZ           & 0.7401 & 0.7420 & 0.7417 & 0.0008 \\
Wafer                          & 0.9927 & 0.9935 & 0.9933 & 0.0003 \\
Wine                           & 0.6296 & 0.6296 & 0.6296 & 0.0000 \\
WordSynonyms                   & 0.6708 & 0.7053 & 0.6984 & 0.0138 \\
Worms                          & 0.7143 & 0.7403 & 0.7299 & 0.0127 \\
WormsTwoClass                  & 0.7403 & 0.7403 & 0.7403 & 0.0000 \\
Yoga                           & 0.8480 & 0.8480 & 0.8480 & 0.0000 \\
\bottomrule
\end{tabular}}
\end{center}

\end{table*}
We list full results for the UCR 2018 repository (minus Crop, as previously mentioned) for both KNN and SVM models trained on 4-window GeoStat representations with two iterations of Laplacian smoothing. Table~\ref{tab:UCR2018Full}. All results were taken over five iterations of cross validation. Of particular interest is the generally small standard deviation over the multiple iterations of cross validation. This is in stark contrast to the much larger variation found in the results of deep learning models in \cite{fawaz2019deep}. 

\begin{table*}[]
\caption{Tables showing accuracy statistics for the UCR 2018 repository with 10 iterations of cross validation. The results listed here are for SVM (left) and KNN (right) trained on 4-window GeoStat representations with two iterations of Laplacian smoothing.}
\label{tab:UCR201810Iter}

\begin{center}
\resizebox*{0.65\textwidth}{0.95\textheight}{
\begin{tabular}{@{}lllll@{}}
\toprule
Dataset (SVM)                          & Min    & Max    & Mean   & Std. Dev  \\
\midrule
ACSF1                          & 0.6200 & 0.6500 & 0.6230 & 0.0090 \\
Adiac                          & 0.7775 & 0.7826 & 0.7780 & 0.0015 \\
AllGestureWiimoteX             & 0.5471 & 0.5471 & 0.5471 & 0.0000 \\
AllGestureWiimoteY             & 0.5557 & 0.5900 & 0.5866 & 0.0103 \\
AllGestureWiimoteZ             & 0.5500 & 0.5500 & 0.5500 & 0.0000 \\
ArrowHead                      & 0.7371 & 0.7371 & 0.7371 & 0.0000 \\
BME                            & 0.9867 & 0.9867 & 0.9867 & 0.0000 \\
Beef                           & 0.7667 & 0.7667 & 0.7667 & 0.0000 \\
BeetleFly                      & 0.5000 & 0.9000 & 0.6750 & 0.1146 \\
BirdChicken                    & 0.9000 & 0.9000 & 0.9000 & 0.0000 \\
CBF                            & 0.9544 & 0.9756 & 0.9734 & 0.0063 \\
Car                            & 0.8333 & 0.8833 & 0.8667 & 0.0211 \\
Chinatown                      & 0.7522 & 0.7522 & 0.7522 & 0.0000 \\
ChlorineConcentration          & 0.6901 & 0.6901 & 0.6901 & 0.0000 \\
CinCECGTorso                   & 0.9246 & 0.9246 & 0.9246 & 0.0000 \\
Coffee                         & 1.0000 & 1.0000 & 1.0000 & 0.0000 \\
Computers                      & 0.5800 & 0.6040 & 0.6012 & 0.0072 \\
CricketX                       & 0.6487 & 0.6487 & 0.6487 & 0.0000 \\
CricketY                       & 0.7000 & 0.7000 & 0.7000 & 0.0000 \\
CricketZ                       & 0.6359 & 0.6846 & 0.6797 & 0.0146 \\
DiatomSizeReduction            & 0.9608 & 0.9608 & 0.9608 & 0.0000 \\
DistalPhalanxOutlineAgeGroup   & 0.7338 & 0.7338 & 0.7338 & 0.0000 \\
DistalPhalanxOutlineCorrect    & 0.7681 & 0.7717 & 0.7710 & 0.0014 \\
DistalPhalanxTW                & 0.6691 & 0.6691 & 0.6691 & 0.0000 \\
DodgerLoopDay                  & 0.3500 & 0.4125 & 0.3625 & 0.0250 \\
DodgerLoopGame                 & 0.7391 & 0.7536 & 0.7522 & 0.0043 \\
DodgerLoopWeekend              & 0.9493 & 0.9710 & 0.9688 & 0.0065 \\
ECG200                         & 0.8100 & 0.8400 & 0.8210 & 0.0137 \\
ECG5000                        & 0.9224 & 0.9362 & 0.9342 & 0.0039 \\
ECGFiveDays                    & 0.9501 & 0.9501 & 0.9501 & 0.0000 \\
EOGHorizontalSignal            & 0.4890 & 0.5028 & 0.4931 & 0.0063 \\
EOGVerticalSignal              & 0.4420 & 0.4448 & 0.4431 & 0.0014 \\
Earthquakes                    & 0.7338 & 0.7482 & 0.7410 & 0.0072 \\
ElectricDevices                & 0.7041 & 0.7041 & 0.7041 & 0.0000 \\
EthanolLevel                   & 0.7580 & 0.7660 & 0.7652 & 0.0024 \\
FaceAll                        & 0.8018 & 0.8018 & 0.8018 & 0.0000 \\
FaceFour                       & 0.7841 & 0.8977 & 0.7955 & 0.0341 \\
FacesUCR                       & 0.8561 & 0.8610 & 0.8595 & 0.0022 \\
FiftyWords                     & 0.7604 & 0.7670 & 0.7624 & 0.0030 \\
Fish                           & 0.9257 & 0.9257 & 0.9257 & 0.0000 \\
FordA                          & 0.8242 & 0.8242 & 0.8242 & 0.0000 \\
FordB                          & 0.6617 & 0.6852 & 0.6758 & 0.0115 \\
FreezerRegularTrain            & 0.9947 & 0.9947 & 0.9947 & 0.0000 \\
FreezerSmallTrain              & 0.9744 & 0.9744 & 0.9744 & 0.0000 \\
Fungi                          & 0.8602 & 0.8602 & 0.8602 & 0.0000 \\
GestureMidAirD1                & 0.6615 & 0.6615 & 0.6615 & 0.0000 \\
GestureMidAirD2                & 0.4923 & 0.5538 & 0.5046 & 0.0246 \\
GestureMidAirD3                & 0.2769 & 0.3231 & 0.3185 & 0.0138 \\
GesturePebbleZ1                & 0.8430 & 0.8837 & 0.8634 & 0.0203 \\
GesturePebbleZ2                & 0.7975 & 0.7975 & 0.7975 & 0.0000 \\
GunPoint                       & 1.0000 & 1.0000 & 1.0000 & 0.0000 \\
GunPointAgeSpan                & 0.9842 & 0.9842 & 0.9842 & 0.0000 \\
GunPointMaleVersusFemale       & 0.9842 & 0.9937 & 0.9889 & 0.0041 \\
GunPointOldVersusYoung         & 1.0000 & 1.0000 & 1.0000 & 0.0000 \\
Ham                            & 0.6571 & 0.6762 & 0.6667 & 0.0095 \\
HandOutlines                   & 0.9324 & 0.9324 & 0.9324 & 0.0000 \\
Haptics                        & 0.4968 & 0.5195 & 0.5081 & 0.0114 \\
Herring                        & 0.6250 & 0.6406 & 0.6359 & 0.0072 \\
HouseTwenty                    & 0.9412 & 0.9664 & 0.9538 & 0.0126 \\
InlineSkate                    & 0.4891 & 0.5291 & 0.5211 & 0.0160 \\
InsectEPGRegularTrain          & 1.0000 & 1.0000 & 1.0000 & 0.0000 \\
InsectEPGSmallTrain            & 0.9719 & 0.9719 & 0.9719 & 0.0000 \\
InsectWingbeatSound            & 0.5354 & 0.5753 & 0.5516 & 0.0193 \\
ItalyPowerDemand               & 0.9378 & 0.9466 & 0.9396 & 0.0035 \\
LargeKitchenAppliances         & 0.7200 & 0.7200 & 0.7200 & 0.0000 \\
Lightning2                     & 0.7377 & 0.7705 & 0.7410 & 0.0098 \\
Lightning7                     & 0.6438 & 0.6712 & 0.6521 & 0.0126 \\
Mallat                         & 0.9390 & 0.9390 & 0.9390 & 0.0000 \\
Meat                           & 0.9500 & 0.9500 & 0.9500 & 0.0000 \\
MedicalImages                  & 0.7092 & 0.7382 & 0.7353 & 0.0087 \\
MelbournePedestrian            & 0.8963 & 0.9008 & 0.8990 & 0.0022 \\
MiddlePhalanxOutlineAgeGroup   & 0.5260 & 0.6234 & 0.6039 & 0.0390 \\
MiddlePhalanxOutlineCorrect    & 0.8351 & 0.8385 & 0.8378 & 0.0014 \\
MiddlePhalanxTW                & 0.5974 & 0.5974 & 0.5974 & 0.0000 \\
MixedShapesRegularTrain        & 0.9559 & 0.9654 & 0.9568 & 0.0028 \\
MixedShapesSmallTrain          & 0.9320 & 0.9320 & 0.9320 & 0.0000 \\
MoteStrain                     & 0.8546 & 0.8546 & 0.8546 & 0.0000 \\
NonInvasiveFetalECGThorax1     & 0.9226 & 0.9252 & 0.9232 & 0.0010 \\
NonInvasiveFetalECGThorax2     & 0.9257 & 0.9262 & 0.9258 & 0.0002 \\
OSULeaf                        & 0.8182 & 0.8182 & 0.8182 & 0.0000 \\
OliveOil                       & 0.9000 & 0.9000 & 0.9000 & 0.0000 \\
PLAID                          & 0.6946 & 0.7095 & 0.6980 & 0.0041 \\
PhalangesOutlinesCorrect       & 0.8077 & 0.8077 & 0.8077 & 0.0000 \\
Phoneme                        & 0.2959 & 0.2959 & 0.2959 & 0.0000 \\
PickupGestureWiimoteZ          & 0.6600 & 0.6600 & 0.6600 & 0.0000 \\
PigAirwayPressure              & 0.1587 & 0.1587 & 0.1587 & 0.0000 \\
PigArtPressure                 & 0.8462 & 0.8462 & 0.8462 & 0.0000 \\
PigCVP                         & 0.4423 & 0.4423 & 0.4423 & 0.0000 \\
Plane                          & 1.0000 & 1.0000 & 1.0000 & 0.0000 \\
PowerCons                      & 0.9389 & 0.9389 & 0.9389 & 0.0000 \\
ProximalPhalanxOutlineAgeGroup & 0.8585 & 0.8585 & 0.8585 & 0.0000 \\
ProximalPhalanxOutlineCorrect  & 0.8454 & 0.8625 & 0.8488 & 0.0069 \\
ProximalPhalanxTW              & 0.7707 & 0.8146 & 0.7795 & 0.0176 \\
RefrigerationDevices           & 0.5013 & 0.5653 & 0.5589 & 0.0192 \\
Rock                           & 0.8000 & 0.8000 & 0.8000 & 0.0000 \\
ScreenType                     & 0.4507 & 0.4507 & 0.4507 & 0.0000 \\
SemgHandGenderCh2              & 0.8833 & 0.9000 & 0.8883 & 0.0076 \\
SemgHandMovementCh2            & 0.7178 & 0.7178 & 0.7178 & 0.0000 \\
SemgHandSubjectCh2             & 0.8222 & 0.8533 & 0.8502 & 0.0093 \\
ShakeGestureWiimoteZ           & 0.8800 & 0.8800 & 0.8800 & 0.0000 \\
ShapeletSim                    & 0.6278 & 0.6500 & 0.6367 & 0.0109 \\
ShapesAll                      & 0.8350 & 0.8350 & 0.8350 & 0.0000 \\
SmallKitchenAppliances         & 0.7920 & 0.7920 & 0.7920 & 0.0000 \\
SmoothSubspace                 & 0.9533 & 0.9600 & 0.9573 & 0.0033 \\
SonyAIBORobotSurface1          & 0.8419 & 0.8419 & 0.8419 & 0.0000 \\
SonyAIBORobotSurface2          & 0.8940 & 0.8940 & 0.8940 & 0.0000 \\
StarLightCurves                & 0.9772 & 0.9772 & 0.9772 & 0.0000 \\
Strawberry                     & 0.9541 & 0.9649 & 0.9638 & 0.0032 \\
SwedishLeaf                    & 0.9584 & 0.9584 & 0.9584 & 0.0000 \\
Symbols                        & 0.9618 & 0.9618 & 0.9618 & 0.0000 \\
SyntheticControl               & 0.9600 & 0.9700 & 0.9610 & 0.0030 \\
ToeSegmentation1               & 0.8509 & 0.8509 & 0.8509 & 0.0000 \\
ToeSegmentation2               & 0.7154 & 0.7769 & 0.7523 & 0.0185 \\
Trace                          & 0.9900 & 0.9900 & 0.9900 & 0.0000 \\
TwoLeadECG                     & 0.9860 & 0.9860 & 0.9860 & 0.0000 \\
TwoPatterns                    & 0.9708 & 0.9708 & 0.9708 & 0.0000 \\
UMD                            & 0.9931 & 0.9931 & 0.9931 & 0.0000 \\
UWaveGestureLibraryAll         & 0.9430 & 0.9497 & 0.9486 & 0.0023 \\
UWaveGestureLibraryX           & 0.8082 & 0.8082 & 0.8082 & 0.0000 \\
UWaveGestureLibraryY           & 0.7457 & 0.7457 & 0.7457 & 0.0000 \\
UWaveGestureLibraryZ           & 0.7440 & 0.7440 & 0.7440 & 0.0000 \\
Wafer                          & 0.9927 & 0.9943 & 0.9929 & 0.0005 \\
Wine                           & 0.5741 & 0.8704 & 0.7222 & 0.1481 \\
WordSynonyms                   & 0.6411 & 0.6897 & 0.6654 & 0.0243 \\
Worms                          & 0.7143 & 0.7143 & 0.7143 & 0.0000 \\
WormsTwoClass                  & 0.8052 & 0.8052 & 0.8052 & 0.0000 \\
Yoga                           & 0.8623 & 0.8623 & 0.8623 & 0.0000 \\
\bottomrule
\end{tabular}
\:\:\:\:\:\:\:\:\:
\begin{tabular}{@{}lllll@{}}
\toprule
Dataset (KNN)                          & Min    & Max    & Mean   & Std. Dev  \\
\midrule
ACSF1                          & 0.7600 & 0.7600 & 0.7600 & 0.0000 \\
Adiac                          & 0.7084 & 0.7366 & 0.7113 & 0.0084 \\
AllGestureWiimoteX             & 0.5800 & 0.5800 & 0.5800 & 0.0000 \\
AllGestureWiimoteY             & 0.5414 & 0.6171 & 0.6096 & 0.0227 \\
AllGestureWiimoteZ             & 0.5971 & 0.5971 & 0.5971 & 0.0000 \\
ArrowHead                      & 0.7886 & 0.7886 & 0.7886 & 0.0000 \\
BME                            & 0.9600 & 0.9600 & 0.9600 & 0.0000 \\
Beef                           & 0.5667 & 0.7333 & 0.5833 & 0.0500 \\
BeetleFly                      & 0.7000 & 0.8000 & 0.7200 & 0.0400 \\
BirdChicken                    & 0.8000 & 0.8000 & 0.8000 & 0.0000 \\
CBF                            & 0.9656 & 0.9800 & 0.9676 & 0.0045 \\
Car                            & 0.7167 & 0.7833 & 0.7583 & 0.0214 \\
Chinatown                      & 0.9242 & 0.9738 & 0.9641 & 0.0134 \\
ChlorineConcentration          & 0.6693 & 0.6693 & 0.6693 & 0.0000 \\
CinCECGTorso                   & 0.8406 & 0.8406 & 0.8406 & 0.0000 \\
Coffee                         & 0.9643 & 1.0000 & 0.9679 & 0.0107 \\
Computers                      & 0.6080 & 0.6480 & 0.6288 & 0.0135 \\
CricketX                       & 0.6231 & 0.6333 & 0.6241 & 0.0031 \\
CricketY                       & 0.6359 & 0.6359 & 0.6359 & 0.0000 \\
CricketZ                       & 0.6282 & 0.6436 & 0.6323 & 0.0045 \\
DiatomSizeReduction            & 0.9379 & 0.9379 & 0.9379 & 0.0000 \\
DistalPhalanxOutlineAgeGroup   & 0.7050 & 0.7194 & 0.7108 & 0.0043 \\
DistalPhalanxOutlineCorrect    & 0.7500 & 0.7500 & 0.7500 & 0.0000 \\
DistalPhalanxTW                & 0.6115 & 0.6547 & 0.6432 & 0.0117 \\
DodgerLoopDay                  & 0.3625 & 0.4500 & 0.4063 & 0.0218 \\
DodgerLoopGame                 & 0.7464 & 0.7464 & 0.7464 & 0.0000 \\
DodgerLoopWeekend              & 0.9493 & 0.9710 & 0.9659 & 0.0086 \\
ECG200                         & 0.8000 & 0.8400 & 0.8080 & 0.0160 \\
ECG5000                        & 0.9413 & 0.9433 & 0.9421 & 0.0010 \\
ECGFiveDays                    & 0.8955 & 0.8955 & 0.8955 & 0.0000 \\
EOGHorizontalSignal            & 0.3453 & 0.4061 & 0.3696 & 0.0184 \\
EOGVerticalSignal              & 0.3177 & 0.3785 & 0.3586 & 0.0154 \\
Earthquakes                    & 0.7338 & 0.7482 & 0.7410 & 0.0064 \\
ElectricDevices                & 0.6631 & 0.6681 & 0.6650 & 0.0024 \\
EthanolLevel                   & 0.4080 & 0.4460 & 0.4246 & 0.0086 \\
FaceAll                        & 0.7550 & 0.8515 & 0.7840 & 0.0442 \\
FaceFour                       & 0.7727 & 0.8409 & 0.7864 & 0.0273 \\
FacesUCR                       & 0.8059 & 0.8420 & 0.8376 & 0.0106 \\
FiftyWords                     & 0.6989 & 0.7187 & 0.7167 & 0.0059 \\
Fish                           & 0.8457 & 0.8743 & 0.8634 & 0.0094 \\
FordA                          & 0.7174 & 0.7318 & 0.7212 & 0.0054 \\
FordB                          & 0.6827 & 0.6889 & 0.6875 & 0.0019 \\
FreezerRegularTrain            & 0.9544 & 0.9740 & 0.9653 & 0.0074 \\
FreezerSmallTrain              & 0.9288 & 0.9309 & 0.9290 & 0.0006 \\
Fungi                          & 0.9409 & 0.9409 & 0.9409 & 0.0000 \\
GestureMidAirD1                & 0.5923 & 0.6231 & 0.6108 & 0.0151 \\
GestureMidAirD2                & 0.5000 & 0.5077 & 0.5069 & 0.0023 \\
GestureMidAirD3                & 0.2385 & 0.3385 & 0.2615 & 0.0284 \\
GesturePebbleZ1                & 0.8430 & 0.8895 & 0.8564 & 0.0116 \\
GesturePebbleZ2                & 0.6519 & 0.7658 & 0.7114 & 0.0257 \\
GunPoint                       & 0.9933 & 0.9933 & 0.9933 & 0.0000 \\
GunPointAgeSpan                & 1.0000 & 1.0000 & 1.0000 & 0.0000 \\
GunPointMaleVersusFemale       & 0.9968 & 0.9968 & 0.9968 & 0.0000 \\
GunPointOldVersusYoung         & 1.0000 & 1.0000 & 1.0000 & 0.0000 \\
Ham                            & 0.5714 & 0.6000 & 0.5952 & 0.0088 \\
HandOutlines                   & 0.8892 & 0.8919 & 0.8900 & 0.0012 \\
Haptics                        & 0.4448 & 0.4740 & 0.4643 & 0.0111 \\
Herring                        & 0.5938 & 0.7500 & 0.6547 & 0.0543 \\
HouseTwenty                    & 0.9244 & 0.9412 & 0.9395 & 0.0050 \\
InlineSkate                    & 0.5782 & 0.5782 & 0.5782 & 0.0000 \\
InsectEPGRegularTrain          & 0.9920 & 0.9920 & 0.9920 & 0.0000 \\
InsectEPGSmallTrain            & 0.9880 & 0.9880 & 0.9880 & 0.0000 \\
InsectWingbeatSound            & 0.4808 & 0.5081 & 0.4957 & 0.0078 \\
ItalyPowerDemand               & 0.9388 & 0.9388 & 0.9388 & 0.0000 \\
LargeKitchenAppliances         & 0.6640 & 0.6960 & 0.6832 & 0.0157 \\
Lightning2                     & 0.7377 & 0.8197 & 0.8115 & 0.0246 \\
Lightning7                     & 0.6712 & 0.6712 & 0.6712 & 0.0000 \\
Mallat                         & 0.9684 & 0.9684 & 0.9684 & 0.0000 \\
Meat                           & 0.9000 & 0.9500 & 0.9433 & 0.0153 \\
MedicalImages                  & 0.6526 & 0.7132 & 0.6950 & 0.0169 \\
MelbournePedestrian            & 0.9147 & 0.9184 & 0.9155 & 0.0015 \\
MiddlePhalanxOutlineAgeGroup   & 0.5455 & 0.5974 & 0.5630 & 0.0219 \\
MiddlePhalanxOutlineCorrect    & 0.7904 & 0.7904 & 0.7904 & 0.0000 \\
MiddlePhalanxTW                & 0.5390 & 0.5779 & 0.5591 & 0.0089 \\
MixedShapesRegularTrain        & 0.9588 & 0.9588 & 0.9588 & 0.0000 \\
MixedShapesSmallTrain          & 0.9134 & 0.9328 & 0.9299 & 0.0062 \\
MoteStrain                     & 0.8578 & 0.8858 & 0.8606 & 0.0084 \\
NonInvasiveFetalECGThorax1     & 0.8656 & 0.8687 & 0.8660 & 0.0009 \\
NonInvasiveFetalECGThorax2     & 0.8707 & 0.8728 & 0.8712 & 0.0008 \\
OSULeaf                        & 0.6983 & 0.8140 & 0.7777 & 0.0337 \\
OliveOil                       & 0.8333 & 0.9333 & 0.8733 & 0.0442 \\
PLAID                          & 0.7486 & 0.7747 & 0.7721 & 0.0078 \\
PhalangesOutlinesCorrect       & 0.7821 & 0.7890 & 0.7876 & 0.0028 \\
Phoneme                        & 0.2126 & 0.2748 & 0.2532 & 0.0232 \\
PickupGestureWiimoteZ          & 0.5400 & 0.7000 & 0.5780 & 0.0610 \\
PigAirwayPressure              & 0.0962 & 0.1731 & 0.1577 & 0.0308 \\
PigArtPressure                 & 0.9038 & 0.9038 & 0.9038 & 0.0000 \\
PigCVP                         & 0.4471 & 0.4471 & 0.4471 & 0.0000 \\
Plane                          & 1.0000 & 1.0000 & 1.0000 & 0.0000 \\
PowerCons                      & 0.8611 & 0.9056 & 0.8972 & 0.0122 \\
ProximalPhalanxOutlineAgeGroup & 0.8390 & 0.8585 & 0.8517 & 0.0076 \\
ProximalPhalanxOutlineCorrect  & 0.8179 & 0.8247 & 0.8199 & 0.0027 \\
ProximalPhalanxTW              & 0.7902 & 0.8098 & 0.8044 & 0.0074 \\
RefrigerationDevices           & 0.4747 & 0.5173 & 0.4875 & 0.0196 \\
Rock                           & 0.6600 & 0.7600 & 0.6900 & 0.0458 \\
ScreenType                     & 0.3760 & 0.4533 & 0.4368 & 0.0239 \\
SemgHandGenderCh2              & 0.9267 & 0.9483 & 0.9332 & 0.0099 \\
SemgHandMovementCh2            & 0.7978 & 0.8000 & 0.7993 & 0.0010 \\
SemgHandSubjectCh2             & 0.8444 & 0.8600 & 0.8544 & 0.0068 \\
ShakeGestureWiimoteZ           & 0.7400 & 0.7800 & 0.7440 & 0.0120 \\
ShapeletSim                    & 0.5833 & 0.6611 & 0.6244 & 0.0302 \\
ShapesAll                      & 0.8433 & 0.8433 & 0.8433 & 0.0000 \\
SmallKitchenAppliances         & 0.7200 & 0.7467 & 0.7413 & 0.0107 \\
SmoothSubspace                 & 0.9467 & 0.9800 & 0.9540 & 0.0131 \\
SonyAIBORobotSurface1          & 0.8602 & 0.8602 & 0.8602 & 0.0000 \\
SonyAIBORobotSurface2          & 0.7754 & 0.8006 & 0.7962 & 0.0075 \\
StarLightCurves                & 0.9720 & 0.9758 & 0.9738 & 0.0014 \\
Strawberry                     & 0.9486 & 0.9541 & 0.9503 & 0.0025 \\
SwedishLeaf                    & 0.9312 & 0.9312 & 0.9312 & 0.0000 \\
Symbols                        & 0.9397 & 0.9397 & 0.9397 & 0.0000 \\
SyntheticControl               & 0.9233 & 0.9233 & 0.9233 & 0.0000 \\
ToeSegmentation1               & 0.6228 & 0.7325 & 0.7154 & 0.0331 \\
ToeSegmentation2               & 0.7538 & 0.8154 & 0.7831 & 0.0145 \\
Trace                          & 1.0000 & 1.0000 & 1.0000 & 0.0000 \\
TwoLeadECG                     & 0.7849 & 0.9298 & 0.8417 & 0.0577 \\
TwoPatterns                    & 0.7553 & 0.7718 & 0.7684 & 0.0045 \\
UMD                            & 0.8958 & 0.9792 & 0.9639 & 0.0265 \\
UWaveGestureLibraryAll         & 0.9196 & 0.9327 & 0.9248 & 0.0064 \\
UWaveGestureLibraryX           & 0.7889 & 0.8102 & 0.7971 & 0.0100 \\
UWaveGestureLibraryY           & 0.7306 & 0.7471 & 0.7433 & 0.0060 \\
UWaveGestureLibraryZ           & 0.7401 & 0.7420 & 0.7415 & 0.0009 \\
Wafer                          & 0.9927 & 0.9935 & 0.9934 & 0.0002 \\
Wine                           & 0.5926 & 0.6296 & 0.6259 & 0.0111 \\
WordSynonyms                   & 0.6708 & 0.7053 & 0.6984 & 0.0138 \\
Worms                          & 0.7143 & 0.7403 & 0.7351 & 0.0104 \\
WormsTwoClass                  & 0.7143 & 0.7403 & 0.7325 & 0.0119 \\
Yoga                           & 0.8480 & 0.8480 & 0.8480 & 0.0000 \\
\bottomrule
\end{tabular}}
\end{center}

\end{table*}
Table~\ref{tab:UCR201810Iter} displays the same results with 10 iterations of cross validation, as is done or the UCR/UEA subset in \cite{fawaz2019deep}. The results listed are consistent with those in Table~\ref{tab:UCR2018Full}.

NOTE FOR REVIEWERS: A sentence in the main text concerning perfect classification for some datasets was meant to be removed in the review draft but mistakenly was left in. As evidenced by Table~\ref{tab:UCR2018Full} and the Supplementary Material, that statement is clearly not true and will be removed in the camera-ready version. We apologize for misleading anyone with its accidental conclusion. All data reported in the Appendix can be found in the Supplementary Material and is directly reproducible by the scripts given.

\subsection{Vessel classification details}
In the following we detail all processing done on the vessel classification dataset. This dataset has a number of challenges, such as large periods of inactivity, that must be dealt with before classifying
\subsubsection{Incorporating sampling rate}
Each vessel trajectory in \cite{GFWData} is sampled nonuniformly and asychronously of one another. The sampling rate for a given vessel is correlated with ocean traffic: if a vessel is around a large number of other vessels, then its sampling rate is more likely to decrease \cite{kroodsma2018tracking}. The number of neighbors of a given vessel (in some radius) was incorporated as a feature used in \cite{kroodsma2018tracking}. Given the lack of data that we have access to compared to \cite{kroodsma2018tracking}, it is not reasonable to expect that such a feature will contain as much information in our case. Nevertheless, the sampling rate may act as a proxy that allows us to incorporate neighbor information. Thus, in addition for the geometric features considered, we also consider distributions based on the sampling rate. More specifically, we consider distributions of lengths of times in which each vessel falls into one of three classifications determined by its speed and sampling rate: active, inactive, and unknown. 

Active time periods are those in which the vessel is moving at a speed greater than or equal to a predefined threshold, whereas inactive time periods are those where the vessel is moving at a speed less than said threshold. Note that it is natural for a vessel to be periodically inactive (e.g. not in use). We specifically chose 0.4 knots as the threshold. For active time periods, all previously mentioned geometric quantities are relevant. For inactive times, the only relevant quantity is the length of period of inactivity; vessels are otherwise almost stationary. 

\begin{figure}
\caption{The time values (in Unix time) of the first timestamps for a trawler. Each unit is one second.}
\label{fig:timestamp}
 \begin{center}
 \centering
\includegraphics[scale=.53]{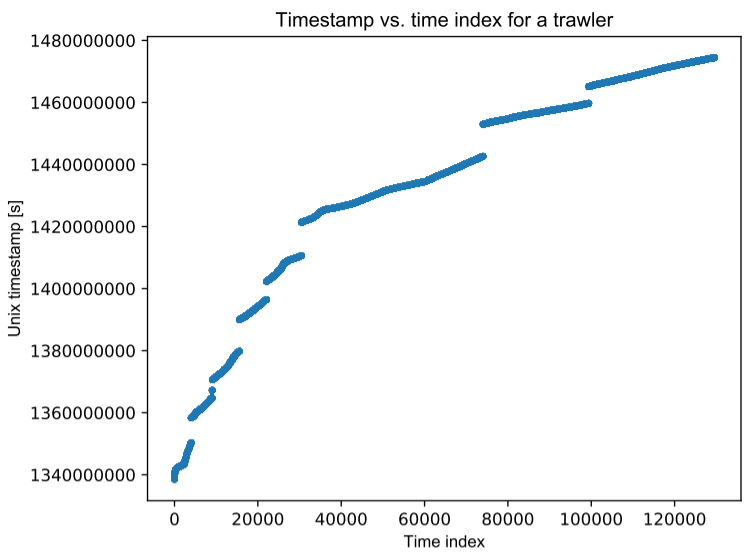}
\end{center}

\end{figure} 

We define an unknown time period to be one in which the amount of time between consecutive samples of a trajectory is at least a minimum threshold; we chose 90 minutes. The length of time between any two samples of a vessel can be quite high; Figure~\ref{fig:timestamp} shows the potential length of gaps for one vessel. Distribution-wise, we consider the distribution of the lengths of time between samples provided that the time between samples is at least 90 minutes. This will act as a proxy for the amount of neighbors a vessel has at any point of time.

\subsubsection{Additional features}
We augment our GeoStat representation a number of other relevant features within our analysis. Two features of interest in this vessel classification problem, distance to shore and distance to port, are given for each trajectory at each time. We also use two variants of the curvature feature defined in the main text: the one proposed, and one obtained without any Laplacian smoothing after differentiating. Using both in tandem gives a multiscale notion of curvature, which initial experiments found important for establishing optimal performance.

\subsubsection{Quantiles}
We employ the following quantiles: 0.001, 0.01, 0.05, 0.1, 0.25, 0.5, 0.75, 0.9, 0.95, 0.99, 0.999.
\subsubsection{Preprocessing}

Not all of the 1258 trajectories available in \cite{GFWData} are immediately usable. A number of these trajectories, for example, do not have a well defined vessel label. Per the supplementary information in \cite{kroodsma2018tracking}, this would suggest a classification of "other fishing'' or "gear/buoy' for each such vessel. Given that the former represents a catch-all category for fishing vessels (for which motion pattern is decidedly unclear) and those in the latter category are not actually vessels, we chose to ignore these in subsequent algorithm development. Note that the performance analysis in \cite{kroodsma2018tracking} did not explicitly consider the gear/buoy class, and achieved approximately 50\% classification accuracy in test, so our decision is not without practical merit. We also chose to remove vessels that did not exhibit any periods of inactivity or any long sampling times; only seven vessels satisfied this property, creating minimal effect. This leaves 1107 trajectories remaining.

Given the length of time that the trajectories span, it is natural that the vessels are motionless for some amount of time. Though the length of such time is of interest, other aspects of the behavior of the vessel are clearly not meaningful. Thus, for each of the 1107 vessels, we first determine if the vessel's speed is fast enough to indicate movement (we chose 0.4 knots as the threshold). 

As previously mentioned, the average such length is itself time dependent. Furthermore, as Figure~\ref{fig:timestamp} showed, the gaps can be quite large. It is not reasonable to assume the behavior before such a gap is similar to that after such a gap. We thus segment the trajectory as follows. We divide each section of the trajectory where the timestamps are consecutively moving or otherwise into subintervals, where two consecutive timestamps are within the same interval if there is at most a 90 minute (5400 second) difference between them. As we are expressly interested in the length between such large time gaps, we record them as they are observed. We keep the intervals (and corresponding portion of the trajectories) if they contain at least four timestamps, and ignore those that have three or fewer timestamps. These segments are ignored because of their tendency to exhibit anomalous behavior. 

Features are derived in two different ways depending on whether the vessel was determined to be actively moving. For the subtrajectories where the vessel was not moving, the only feature of interest is the length of time of such an event. We record this by looking at the difference between the first and last timestamp, and then ignore the segment for the remainder of our analysis.

For the subtrajectories where the vessel is in motion, we first perform piecewise linear interpolation along the time interval in which the subtrajectory is defined. Unlike for the univariate setting, we employ ten iterations of Laplacian smoothing to smoothen our data; this was chosen empirically to remove all possible noise that might have had negative effects in subsequent analysis.

We then constructed samples for each distribution of geometric features by extracting the average value of the interpolated quantities over a ten minute window. We also included the average values for portions of subtrajectories less than ten minutes (which only occurs towards the end values) as samples for our distributions. Samples for values of distributions related to sampling rate are computed in the obvious way: measure the relevant lengths of time.

All features are Z-normalized.

\subsubsection{Confusion matrices}

\begin{figure*}
\caption{Confusion matrices for one 10-fold nested cross validation experiment on the vessel classification dataset. Left: KNN, Right: SVM}\label{fig:Confusion}
    \begin{center}

        \includegraphics[scale=0.6]{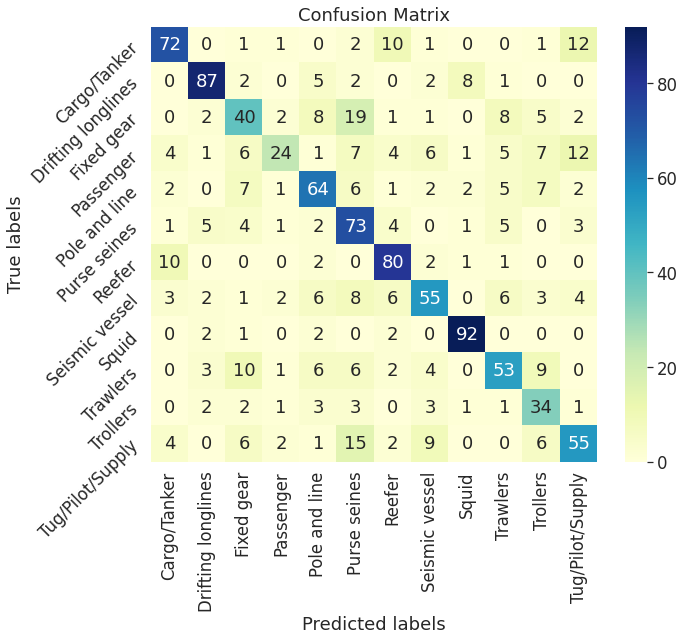}
        \hfill
        \includegraphics[scale=0.6]{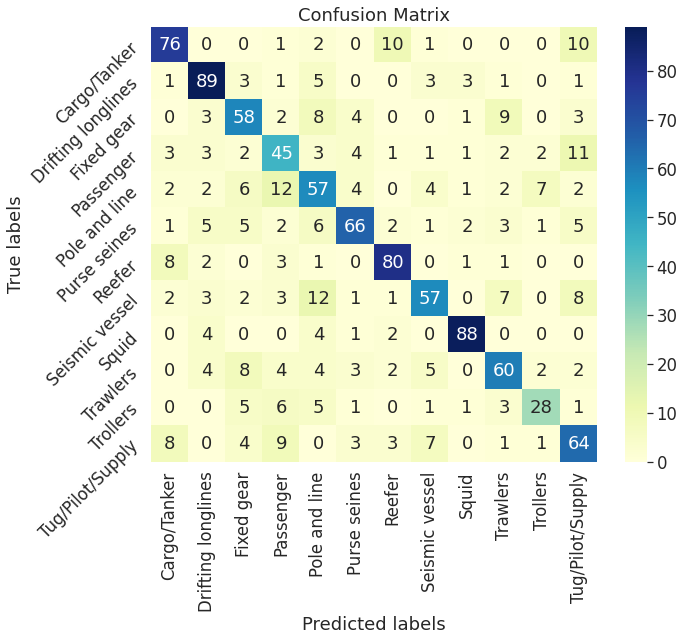}

    \end{center}
\end{figure*}

In Figure \ref{fig:Confusion}, we show the confusion matrices for both KNN and SVM for a fixed, illustrative run of 10-fold nested cross validation. Looking at the confusion matrices, we see that perhaps some mislabeling is warranted. In particular, we note three main types of mislabeling: fishing vessels getting labeled as other types of fishing vessels, reefers (refrigerated cargo) and cargo vessels getting confused, and tugboats getting confused with various classes. It is reasonable to expect that different types of fishing vessels may exhibit similar movement patterns; the same is true for reefers and cargo vessels. High confusion with tug/pilot/supply is also interesting but not surprising; a ship broken down at sea that requires a tugboat is likely to exhibit similar movement patterns to a tugboat.

\begin{figure*}
\caption{Confusion matrices for one 10-fold nested cross validation experiment on the vessel classification dataset for the binary problem of whether a vessel is involved in fishing. Left: KNN, Right: SVM}\label{fig:ConfusionFishing}
    \begin{center}

        \includegraphics[scale=0.55]{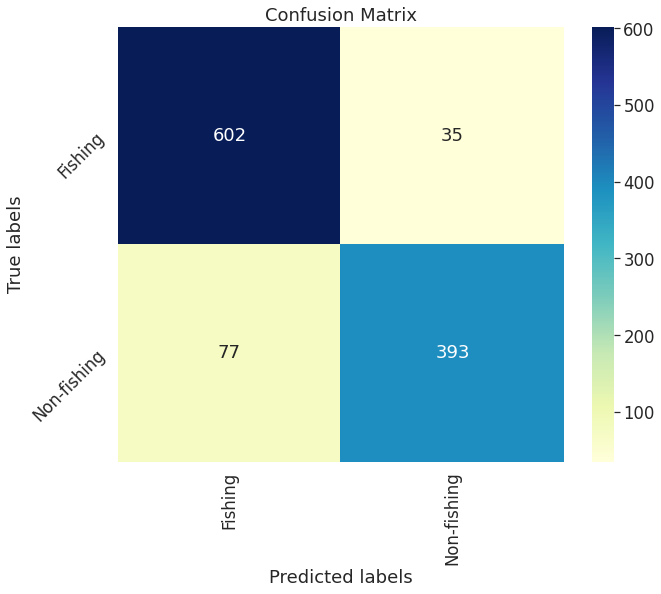}
        \hfill
        \includegraphics[scale=0.55]{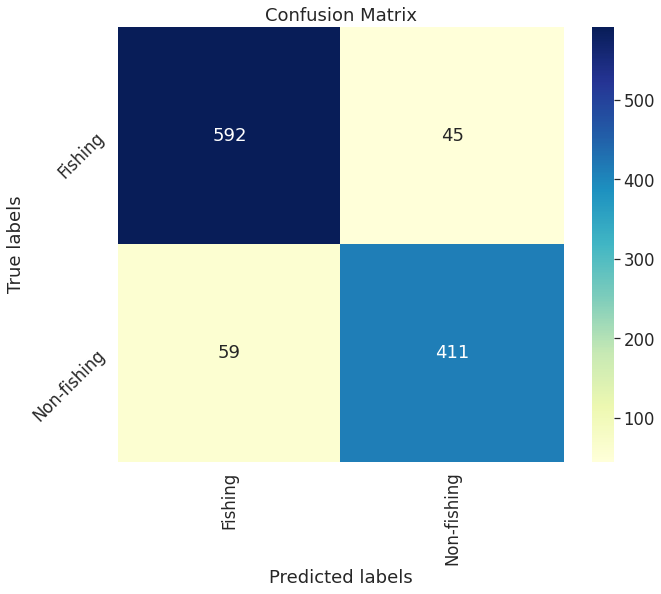}

    \end{center}
\end{figure*}

To give some credence to our claim that mislabeling of one fishing vessel with another is reasonable, we also show in Figure~\ref{fig:ConfusionFishing} the confusion matrices for the binary classification problem of whether a vessel is involved in fishing. Here, the KNN model achieves an accuracy of 89.88\% while the SVM model achieves an accuracy of 90.61\%. Again, note that this is with a nested cross validation procedure on approximately two percent of the data available in \cite{kroodsma2018tracking}, the model in which achieves a near-perfect accuracy score. This further supports that our features are accurately capturing quantities of interest.

\subsection{Details of empirical studies}
\subsubsection{Individual windows}
\begin{table*}[]
\caption{Tables showing accuracy statistics for the UCR 2018 repository for models trained on the individual windows of a 6-window GeoStat representation.}
\label{tab:IndWindow}

\begin{center}
\resizebox*{0.95\textwidth}{0.95\textheight}{
\begin{tabular}{@{}lllllll@{}}
\toprule

Dataset (SVM)                & Window 1 & Window 2 & Window 3 & Window 4 & Window 5 & Window 6 \\
\midrule
ACSF1                        & 0.6000   & 0.4860   & 0.4700   & 0.5800   & 0.5720   & 0.7200   \\
Adiac                        & 0.6552   & 0.5402   & 0.6496   & 0.6486   & 0.5995   & 0.6547   \\
AllGestureWiimoteX           & 0.2486   & 0.3543   & 0.3871   & 0.3991   & 0.3371   & 0.3069   \\
AllGestureWiimoteY           & 0.3271   & 0.3383   & 0.3786   & 0.3686   & 0.3251   & 0.3357   \\
AllGestureWiimoteZ           & 0.2929   & 0.3037   & 0.3571   & 0.3429   & 0.3171   & 0.3080   \\
ArrowHead                    & 0.6000   & 0.7109   & 0.7143   & 0.7200   & 0.6377   & 0.5943   \\
BME                          & 0.6533   & 0.4413   & 0.3773   & 0.4293   & 0.5267   & 0.6747   \\
Beef                         & 0.7333   & 0.5333   & 0.6867   & 0.5800   & 0.4667   & 0.6133   \\
BeetleFly                    & 0.6900   & 0.8000   & 0.6500   & 0.6500   & 0.7500   & 0.8400   \\
BirdChicken                  & 0.7600   & 0.7000   & 0.9000   & 0.9500   & 0.7000   & 0.9500   \\
CBF                          & 0.5756   & 0.8522   & 0.7289   & 0.7222   & 0.6262   & 0.5467   \\
Car                          & 0.6833   & 0.6700   & 0.7167   & 0.6567   & 0.5633   & 0.5100   \\
Chinatown                    & 0.9125   & 0.8058   & 0.7114   & 0.7977   & 0.7743   & 0.5090   \\
ChlorineConcentration        & 0.5953   & 0.6461   & 0.6089   & 0.5916   & 0.6883   & 0.6133   \\
CinCECGTorso                 & 0.5970   & 0.7533   & 0.6951   & 0.8275   & 0.7448   & 0.6197   \\
Coffee                       & 0.9143   & 0.8357   & 0.8071   & 0.9286   & 1.0000   & 0.8929   \\
Computers                    & 0.6624   & 0.7120   & 0.7040   & 0.6960   & 0.6440   & 0.6272   \\
CricketX                     & 0.2641   & 0.5256   & 0.4692   & 0.3590   & 0.2410   & 0.1487   \\
CricketY                     & 0.3949   & 0.5538   & 0.4462   & 0.3241   & 0.2774   & 0.2200   \\
CricketZ                     & 0.2641   & 0.5462   & 0.4821   & 0.3179   & 0.2308   & 0.1564   \\
DiatomSizeReduction          & 0.9412   & 0.8431   & 0.9222   & 0.8856   & 0.8595   & 0.7712   \\
DistalPhalanxOutlineAgeGroup & 0.6691   & 0.6719   & 0.6835   & 0.7050   & 0.7554   & 0.7338   \\
DistalPhalanxOutlineCorrect  & 0.7377   & 0.6710   & 0.7464   & 0.7464   & 0.6884   & 0.6681   \\
DistalPhalanxTW              & 0.6129   & 0.5741   & 0.5770   & 0.6691   & 0.6676   & 0.6878   \\
DodgerLoopDay                & 0.2525   & 0.3950   & 0.3300   & 0.2450   & 0.3275   & 0.2850   \\
DodgerLoopGame               & 0.4783   & 0.5232   & 0.5000   & 0.4855   & 0.5783   & 0.8333   \\
DodgerLoopWeekend            & 0.8768   & 0.9638   & 0.9638   & 0.6725   & 0.7652   & 0.5899   \\
ECG200                       & 0.8180   & 0.8120   & 0.8300   & 0.8000   & 0.8100   & 0.7700   \\
ECG5000                      & 0.9251   & 0.9139   & 0.8451   & 0.8269   & 0.9254   & 0.9352   \\
ECGFiveDays                  & 0.6202   & 0.6002   & 0.9377   & 0.7271   & 0.7909   & 0.9782   \\
EOGHorizontalSignal          & 0.0773   & 0.2044   & 0.3702   & 0.2155   & 0.2039   & 0.2315   \\
EOGVerticalSignal            & 0.0978   & 0.2514   & 0.2569   & 0.2105   & 0.1773   & 0.1298   \\
Earthquakes                  & 0.7482   & 0.7482   & 0.7482   & 0.7482   & 0.7482   & 0.7396   \\
EthanolLevel                 & 0.3064   & 0.3236   & 0.2784   & 0.2660   & 0.7372   & 0.5188   \\
FaceAll                      & 0.7763   & 0.6207   & 0.6320   & 0.5231   & 0.5414   & 0.4592   \\
FaceFour                     & 0.6886   & 0.7386   & 0.7159   & 0.7886   & 0.5295   & 0.3295   \\
FacesUCR                     & 0.4513   & 0.5971   & 0.5520   & 0.7102   & 0.6042   & 0.6112   \\
FiftyWords                   & 0.5068   & 0.4760   & 0.4681   & 0.4703   & 0.4923   & 0.4681   \\
Fish                         & 0.7749   & 0.7829   & 0.7463   & 0.7440   & 0.6640   & 0.3783   \\
FordA                        & 0.7583   & 0.7395   & 0.7644   & 0.7267   & 0.7414   & 0.7273   \\
FordB                        & 0.6210   & 0.6464   & 0.6457   & 0.6432   & 0.6148   & 0.6531   \\
FreezerRegularTrain          & 0.9505   & 0.9933   & 0.8652   & 0.8747   & 0.8340   & 0.7932   \\
FreezerSmallTrain            & 0.8902   & 0.9926   & 0.8920   & 0.8677   & 0.7177   & 0.7888   \\
Fungi                        & 0.1237   & 0.4570   & 0.7151   & 0.6667   & 0.3495   & 0.2849   \\
GestureMidAirD1              & 0.2200   & 0.3892   & 0.3415   & 0.3400   & 0.3431   & 0.3031   \\
GestureMidAirD2              & 0.2262   & 0.2846   & 0.2615   & 0.2600   & 0.2385   & 0.2277   \\
GestureMidAirD3              & 0.1231   & 0.1754   & 0.1831   & 0.1646   & 0.1338   & 0.1754   \\
GesturePebbleZ1              & 0.3442   & 0.5651   & 0.6395   & 0.6628   & 0.8384   & 0.6826   \\
GesturePebbleZ2              & 0.3038   & 0.4975   & 0.5633   & 0.6114   & 0.7089   & 0.6481   \\
GunPoint                     & 0.6933   & 0.9400   & 0.7293   & 0.8173   & 0.9147   & 0.8467   \\
GunPointAgeSpan              & 0.8741   & 0.9462   & 0.8703   & 0.8184   & 0.9253   & 0.8513   \\
GunPointMaleVersusFemale     & 0.9272   & 0.9190   & 0.9652   & 0.9127   & 0.8468   & 0.8544   \\
GunPointOldVersusYoung       & 0.9714   & 0.9740   & 0.9937   & 1.0000   & 0.9683   & 0.9663   \\
Ham                          & 0.6686   & 0.5524   & 0.6629   & 0.5257   & 0.6286   & 0.6419   \\
HandOutlines                 & 0.7146   & 0.9141   & 0.8486   & 0.7724   & 0.7405   & 0.6784   \\
Haptics                      & 0.3805   & 0.4123   & 0.3727   & 0.3831   & 0.3955   & 0.4312   \\
Herring                      & 0.6594   & 0.6281   & 0.5938   & 0.5938   & 0.5938   & 0.5281   \\
HouseTwenty                  & 0.9076   & 0.8941   & 0.9092   & 0.8252   & 0.6168   & 0.6571   \\
InlineSkate                  & 0.3982   & 0.2524   & 0.2727   & 0.2687   & 0.3131   & 0.4098   \\
InsectEPGRegularTrain        & 1.0000   & 0.9839   & 0.9823   & 0.9807   & 0.9960   & 1.0000   \\
InsectEPGSmallTrain          & 0.9157   & 1.0000   & 0.8835   & 0.9687   & 0.9558   & 0.9839   \\
InsectWingbeatSound          & 0.4725   & 0.4313   & 0.4626   & 0.3715   & 0.3424   & 0.3104   \\
ItalyPowerDemand             & 0.6200   & 0.8140   & 0.7794   & 0.8185   & 0.9504   & 0.9252   \\
LargeKitchenAppliances       & 0.5136   & 0.5755   & 0.6133   & 0.5947   & 0.6027   & 0.5243   \\
Lightning2                   & 0.6164   & 0.5803   & 0.7344   & 0.6361   & 0.5410   & 0.6361   \\
Lightning7                   & 0.4164   & 0.5315   & 0.4849   & 0.5562   & 0.6137   & 0.5342   \\
Mallat                       & 0.5850   & 0.6493   & 0.5082   & 0.6098   & 0.6235   & 0.7224   \\
Meat                         & 0.9300   & 0.8667   & 0.9867   & 0.8300   & 0.9133   & 0.8000   \\
MedicalImages                & 0.7271   & 0.5382   & 0.5566   & 0.5487   & 0.5737   & 0.5755   \\
MelbournePedestrian          & 0.6103   & 0.8163   & 0.8115   & 0.7708   & 0.8115   & 0.6134   \\
MiddlePhalanxOutlineAgeGroup & 0.6039   & 0.5766   & 0.5909   & 0.5753   & 0.5792   & 0.6182   \\
MiddlePhalanxOutlineCorrect  & 0.6564   & 0.6460   & 0.5828   & 0.7237   & 0.7443   & 0.7113   \\
MiddlePhalanxTW              & 0.5455   & 0.5468   & 0.5974   & 0.5779   & 0.5610   & 0.5831   \\
MixedShapesRegularTrain      & 0.9010   & 0.9131   & 0.8973   & 0.8942   & 0.9015   & 0.8907   \\
MixedShapesSmallTrain        & 0.8246   & 0.8456   & 0.8355   & 0.8284   & 0.8285   & 0.8304   \\
MoteStrain                   & 0.8858   & 0.7663   & 0.6414   & 0.8043   & 0.8227   & 0.7534   \\
NonInvasiveFetalECGThorax1   & 0.8588   & 0.5069   & 0.5954   & 0.3968   & 0.5812   & 0.7235   \\
NonInvasiveFetalECGThorax2   & 0.8744   & 0.5348   & 0.6651   & 0.4928   & 0.6590   & 0.8132   \\
OSULeaf                      & 0.6405   & 0.6636   & 0.6372   & 0.6496   & 0.6802   & 0.7107   \\
OliveOil                     & 0.7867   & 0.9000   & 0.8600   & 0.8667   & 0.8000   & 0.7000   \\
PLAID                        & 0.5624   & 0.4317   & 0.4510   & 0.4872   & 0.4860   & 0.4935   \\
PhalangesOutlinesCorrect     & 0.7086   & 0.6993   & 0.7133   & 0.7664   & 0.6900   & 0.6876   \\
Phoneme                      & 0.1398   & 0.2120   & 0.1963   & 0.1755   & 0.1685   & 0.1558   \\
PickupGestureWiimoteZ        & 0.4000   & 0.4840   & 0.4040   & 0.4240   & 0.4960   & 0.2960   \\
PigAirwayPressure            & 0.1106   & 0.1058   & 0.1154   & 0.0808   & 0.1250   & 0.1231   \\
PigArtPressure               & 0.7644   & 0.6990   & 0.7212   & 0.6827   & 0.6846   & 0.6808   \\
PigCVP                       & 0.2163   & 0.2308   & 0.2644   & 0.2067   & 0.2788   & 0.1731   \\
Plane                        & 0.9905   & 0.9905   & 1.0000   & 1.0000   & 0.9733   & 0.9790   \\
PowerCons                    & 0.6967   & 0.7600   & 0.7522   & 0.8000   & 0.9900   & 0.8833   \\
RefrigerationDevices         & 0.5200   & 0.4976   & 0.4613   & 0.4635   & 0.4773   & 0.4960   \\
Rock                         & 0.6200   & 0.5200   & 0.5040   & 0.5400   & 0.7280   & 0.7800   \\
ScreenType                   & 0.3413   & 0.3600   & 0.3515   & 0.4747   & 0.4533   & 0.4720   \\
SemgHandGenderCh2            & 0.7107   & 0.7807   & 0.6697   & 0.7637   & 0.6700   & 0.6587   \\
SemgHandMovementCh2          & 0.4627   & 0.4244   & 0.3578   & 0.3600   & 0.3831   & 0.3733   \\
SemgHandSubjectCh2           & 0.5071   & 0.5067   & 0.4507   & 0.4938   & 0.4311   & 0.4173   \\
ShakeGestureWiimoteZ         & 0.4000   & 0.6200   & 0.5920   & 0.5120   & 0.5680   & 0.5440   \\
ShapeletSim                  & 0.5556   & 0.5267   & 0.5222   & 0.4500   & 0.5444   & 0.5267   \\
ShapesAll                    & 0.6450   & 0.6500   & 0.5867   & 0.6350   & 0.6517   & 0.6600   \\
SmallKitchenAppliances       & 0.7637   & 0.7483   & 0.7397   & 0.7227   & 0.7157   & 0.7093   \\
SmoothSubspace               & 0.5987   & 0.7267   & 0.6547   & 0.7800   & 0.6280   & 0.6067   \\
SonyAIBORobotSurface1        & 0.8802   & 0.7687   & 0.5324   & 0.5584   & 0.7278   & 0.7304   \\
SonyAIBORobotSurface2        & 0.7400   & 0.8233   & 0.6751   & 0.7209   & 0.8361   & 0.7933   \\
StarLightCurves              & 0.9530   & 0.9517   & 0.9149   & 0.9451   & 0.9329   & 0.9381   \\
Strawberry                   & 0.9054   & 0.9486   & 0.9405   & 0.9486   & 0.8465   & 0.8865   \\
SwedishLeaf                  & 0.8480   & 0.7488   & 0.7952   & 0.7904   & 0.7514   & 0.8400   \\
Symbols                      & 0.6834   & 0.8344   & 0.8133   & 0.8683   & 0.8613   & 0.7833   \\
SyntheticControl             & 0.8500   & 0.7527   & 0.8000   & 0.8467   & 0.6693   & 0.8333   \\
ToeSegmentation1             & 0.5237   & 0.6772   & 0.5860   & 0.6588   & 0.7061   & 0.6746   \\
ToeSegmentation2             & 0.6969   & 0.6831   & 0.6292   & 0.5692   & 0.6462   & 0.5015   \\
Trace                        & 0.6500   & 0.6880   & 0.6600   & 0.7620   & 0.8100   & 0.5000   \\
TwoLeadECG                   & 0.7073   & 0.8999   & 0.9377   & 0.7858   & 0.7208   & 0.6653   \\
TwoPatterns                  & 0.3305   & 0.3895   & 0.4090   & 0.4357   & 0.5172   & 0.3995   \\
UMD                          & 0.6667   & 0.6417   & 0.5153   & 0.4653   & 0.5750   & 0.6042   \\
UWaveGestureLibraryAll       & 0.7462   & 0.8143   & 0.7024   & 0.7083   & 0.7151   & 0.6493   \\
UWaveGestureLibraryX         & 0.5168   & 0.6499   & 0.5952   & 0.5931   & 0.6200   & 0.5823   \\
UWaveGestureLibraryY         & 0.4963   & 0.5860   & 0.5324   & 0.5338   & 0.5357   & 0.5248   \\
UWaveGestureLibraryZ         & 0.5059   & 0.6228   & 0.5461   & 0.5561   & 0.5353   & 0.4997   \\
Wafer                        & 0.9924   & 0.9869   & 0.9401   & 0.9518   & 0.9839   & 0.9937   \\
Wine                         & 0.7481   & 0.6852   & 0.7889   & 0.6852   & 0.5000   & 0.5556   \\
WordSynonyms                 & 0.4326   & 0.4295   & 0.4075   & 0.4279   & 0.4577   & 0.4636   \\
Worms                        & 0.5870   & 0.5714   & 0.6883   & 0.5714   & 0.5455   & 0.6364   \\
WormsTwoClass                & 0.6883   & 0.6286   & 0.7584   & 0.7273   & 0.7299   & 0.6701   \\
Yoga                         & 0.7713   & 0.7257   & 0.6193   & 0.7103   & 0.7407   & 0.8083  \\
\bottomrule
\end{tabular}
\:\:\:\:\:\:\:\:\:
\begin{tabular}{@{}lllllll@{}}
\toprule
Dataset (KNN)                & Window 1 & Window 2 & Window 3 & Window 4 & Window 5 & Window 6 \\
\midrule
ACSF1                        & 0.6160 & 0.4700 & 0.5900 & 0.6000 & 0.5600 & 0.7900 \\
Adiac                        & 0.5499 & 0.4972 & 0.5703 & 0.5678 & 0.5243 & 0.5473 \\
AllGestureWiimoteX           & 0.2357 & 0.3414 & 0.3877 & 0.3757 & 0.3223 & 0.2780 \\
AllGestureWiimoteY           & 0.2709 & 0.3489 & 0.3971 & 0.3720 & 0.3351 & 0.3429 \\
AllGestureWiimoteZ           & 0.2966 & 0.3077 & 0.3471 & 0.3780 & 0.3300 & 0.2883 \\
ArrowHead                    & 0.5486 & 0.6971 & 0.7017 & 0.6903 & 0.5886 & 0.5829 \\
BME                          & 0.6973 & 0.4053 & 0.4400 & 0.4227 & 0.4653 & 0.6200 \\
Beef                         & 0.6800 & 0.3800 & 0.4867 & 0.5000 & 0.4933 & 0.5067 \\
BeetleFly                    & 0.7000 & 0.7000 & 0.5900 & 0.6000 & 0.8500 & 0.6600 \\
BirdChicken                  & 0.8500 & 0.8000 & 0.8500 & 0.7900 & 0.7000 & 0.8500 \\
CBF                          & 0.5758 & 0.8311 & 0.6576 & 0.7347 & 0.5989 & 0.5678 \\
Car                          & 0.7167 & 0.5367 & 0.6767 & 0.6567 & 0.5267 & 0.4800 \\
Chinatown                    & 0.9662 & 0.8484 & 0.7965 & 0.8222 & 0.6880 & 0.6636 \\
ChlorineConcentration        & 0.5690 & 0.6199 & 0.5923 & 0.5698 & 0.6963 & 0.5968 \\
CinCECGTorso                 & 0.5732 & 0.6920 & 0.7428 & 0.8116 & 0.7122 & 0.6510 \\
Coffee                       & 0.7714 & 0.7857 & 0.7071 & 0.9571 & 1.0000 & 0.7929 \\
Computers                    & 0.6816 & 0.7008 & 0.7120 & 0.6664 & 0.6776 & 0.6496 \\
CricketX                     & 0.2641 & 0.5164 & 0.4092 & 0.3210 & 0.2082 & 0.1554 \\
CricketY                     & 0.3656 & 0.5615 & 0.4041 & 0.3549 & 0.2826 & 0.2221 \\
CricketZ                     & 0.2595 & 0.5621 & 0.4349 & 0.3672 & 0.2036 & 0.1405 \\
DiatomSizeReduction          & 0.8922 & 0.7190 & 0.9118 & 0.8824 & 0.8268 & 0.7974 \\
DistalPhalanxOutlineAgeGroup & 0.6662 & 0.6590 & 0.6504 & 0.7137 & 0.7194 & 0.7122 \\
DistalPhalanxOutlineCorrect  & 0.6572 & 0.6464 & 0.6986 & 0.7420 & 0.6551 & 0.6652 \\
DistalPhalanxTW              & 0.5568 & 0.5554 & 0.5424 & 0.6374 & 0.6763 & 0.6388 \\
DodgerLoopDay                & 0.2750 & 0.3750 & 0.3775 & 0.2375 & 0.2550 & 0.1750 \\
DodgerLoopGame               & 0.5145 & 0.4913 & 0.4986 & 0.5217 & 0.5638 & 0.7899 \\
DodgerLoopWeekend            & 0.8841 & 0.9783 & 0.9275 & 0.8058 & 0.7667 & 0.5710 \\
ECG200                       & 0.8400 & 0.8200 & 0.8300 & 0.7700 & 0.7280 & 0.7340 \\
ECG5000                      & 0.9268 & 0.9054 & 0.8063 & 0.8179 & 0.9220 & 0.9319 \\
ECGFiveDays                  & 0.6581 & 0.6490 & 0.8583 & 0.7415 & 0.7482 & 0.9129 \\
EOGHorizontalSignal          & 0.0840 & 0.1906 & 0.3315 & 0.1994 & 0.2105 & 0.1796 \\
EOGVerticalSignal            & 0.0657 & 0.2713 & 0.2414 & 0.1873 & 0.1282 & 0.1331 \\
Earthquakes                  & 0.7468 & 0.7410 & 0.7482 & 0.7410 & 0.7381 & 0.7511 \\
EthanolLevel                 & 0.2672 & 0.2980 & 0.3068 & 0.2484 & 0.5784 & 0.4168 \\
FaceAll                      & 0.6904 & 0.5781 & 0.6080 & 0.4770 & 0.4712 & 0.4021 \\
FaceFour                     & 0.5841 & 0.8068 & 0.7023 & 0.7409 & 0.4568 & 0.2841 \\
FacesUCR                     & 0.4126 & 0.5317 & 0.5307 & 0.6659 & 0.5433 & 0.5412 \\
FiftyWords                   & 0.4743 & 0.4642 & 0.4488 & 0.4440 & 0.4681 & 0.4602 \\
Fish                         & 0.7131 & 0.7714 & 0.7589 & 0.7486 & 0.5920 & 0.3749 \\
FordA                        & 0.6997 & 0.6950 & 0.7317 & 0.6974 & 0.7009 & 0.7029 \\
FordB                        & 0.6160 & 0.6412 & 0.6311 & 0.6504 & 0.6267 & 0.6299 \\
FreezerRegularTrain          & 0.9502 & 0.9811 & 0.8722 & 0.8778 & 0.8512 & 0.7908 \\
FreezerSmallTrain            & 0.8846 & 0.9639 & 0.9052 & 0.8606 & 0.7446 & 0.7182 \\
Fungi                        & 0.1344 & 0.4624 & 0.8065 & 0.7258 & 0.4194 & 0.3011 \\
GestureMidAirD1              & 0.2308 & 0.3077 & 0.2738 & 0.3338 & 0.2892 & 0.2846 \\
GestureMidAirD2              & 0.2169 & 0.2492 & 0.2385 & 0.2523 & 0.2400 & 0.2154 \\
GestureMidAirD3              & 0.1231 & 0.2000 & 0.1646 & 0.1538 & 0.1585 & 0.1908 \\
GesturePebbleZ1              & 0.3163 & 0.4198 & 0.5756 & 0.6372 & 0.7698 & 0.5640 \\
GesturePebbleZ2              & 0.1949 & 0.3899 & 0.4380 & 0.5481 & 0.5696 & 0.5582 \\
GunPoint                     & 0.7693 & 0.9733 & 0.7627 & 0.8147 & 0.9067 & 0.8480 \\
GunPointAgeSpan              & 0.8734 & 0.9772 & 0.8715 & 0.8576 & 0.9114 & 0.8778 \\
GunPointMaleVersusFemale     & 0.9500 & 0.9563 & 0.9810 & 0.9329 & 0.8766 & 0.8785 \\
GunPointOldVersusYoung       & 0.9841 & 0.9873 & 0.9905 & 0.9968 & 0.9778 & 0.9714 \\
Ham                          & 0.6457 & 0.5048 & 0.5429 & 0.5124 & 0.5981 & 0.4533 \\
HandOutlines                 & 0.7076 & 0.8443 & 0.7832 & 0.7578 & 0.7303 & 0.6422 \\
Haptics                      & 0.3558 & 0.3747 & 0.3714 & 0.3435 & 0.3948 & 0.4019 \\
Herring                      & 0.6094 & 0.5781 & 0.5719 & 0.5750 & 0.5844 & 0.5500 \\
HouseTwenty                  & 0.9328 & 0.9580 & 0.9076 & 0.8538 & 0.7513 & 0.6571 \\
InlineSkate                  & 0.4149 & 0.2658 & 0.2633 & 0.2945 & 0.3018 & 0.3742 \\
InsectEPGRegularTrain        & 0.9454 & 0.9839 & 0.9783 & 0.9703 & 0.9703 & 0.9454 \\
InsectEPGSmallTrain          & 0.9357 & 0.9116 & 0.9518 & 0.8618 & 0.9558 & 0.8353 \\
InsectWingbeatSound          & 0.4468 & 0.4086 & 0.4016 & 0.3270 & 0.3102 & 0.2735 \\
ItalyPowerDemand             & 0.6064 & 0.8091 & 0.8016 & 0.8148 & 0.9460 & 0.9333 \\
LargeKitchenAppliances       & 0.5147 & 0.5083 & 0.6272 & 0.5947 & 0.5563 & 0.5733 \\
Lightning2                   & 0.5803 & 0.6131 & 0.7344 & 0.6393 & 0.7607 & 0.7180 \\
Lightning7                   & 0.4493 & 0.4219 & 0.4384 & 0.4932 & 0.6384 & 0.5041 \\
Mallat                       & 0.5267 & 0.6404 & 0.4289 & 0.5976 & 0.6792 & 0.6398 \\
Meat                         & 0.8933 & 0.8600 & 0.9967 & 0.8733 & 0.8967 & 0.8167 \\
MedicalImages                & 0.6971 & 0.5421 & 0.5429 & 0.5316 & 0.5611 & 0.5776 \\
MelbournePedestrian          & 0.5550 & 0.8073 & 0.8315 & 0.7847 & 0.7966 & 0.6113 \\
MiddlePhalanxOutlineAgeGroup & 0.5662 & 0.5455 & 0.5909 & 0.5545 & 0.6078 & 0.5584 \\
MiddlePhalanxOutlineCorrect  & 0.6275 & 0.6742 & 0.5759 & 0.6866 & 0.7478 & 0.7182 \\
MiddlePhalanxTW              & 0.5195 & 0.5052 & 0.5571 & 0.5623 & 0.5364 & 0.5286 \\
MixedShapesRegularTrain      & 0.8831 & 0.9002 & 0.9037 & 0.8859 & 0.8721 & 0.8854 \\
MixedShapesSmallTrain        & 0.7946 & 0.8130 & 0.8407 & 0.8099 & 0.8045 & 0.8118 \\
MoteStrain                   & 0.7542 & 0.8075 & 0.7196 & 0.7821 & 0.8395 & 0.7139 \\
NonInvasiveFetalECGThorax1   & 0.7838 & 0.4208 & 0.4731 & 0.3351 & 0.4743 & 0.5999 \\
NonInvasiveFetalECGThorax2   & 0.8490 & 0.4515 & 0.5410 & 0.4083 & 0.5641 & 0.7359 \\
OSULeaf                      & 0.6331 & 0.6025 & 0.6033 & 0.5711 & 0.6083 & 0.6256 \\
OliveOil                     & 0.7867 & 0.8467 & 0.7067 & 0.9133 & 0.8933 & 0.7000 \\
PLAID                        & 0.5996 & 0.5456 & 0.5903 & 0.5810 & 0.6387 & 0.5989 \\
PhalangesOutlinesCorrect     & 0.6825 & 0.6811 & 0.6900 & 0.7354 & 0.6744 & 0.6783 \\
Phoneme                      & 0.1516 & 0.1646 & 0.1590 & 0.1826 & 0.1536 & 0.1373 \\
PickupGestureWiimoteZ        & 0.4200 & 0.4400 & 0.3240 & 0.3560 & 0.3920 & 0.2960 \\
PigAirwayPressure            & 0.1221 & 0.1144 & 0.1106 & 0.0817 & 0.1058 & 0.1250 \\
PigArtPressure               & 0.7548 & 0.6875 & 0.7404 & 0.7115 & 0.7106 & 0.7067 \\
PigCVP                       & 0.2404 & 0.2837 & 0.2644 & 0.1971 & 0.3317 & 0.2212 \\
Plane                        & 0.9600 & 0.9810 & 0.9905 & 0.9905 & 0.9810 & 0.9714 \\
PowerCons                    & 0.7411 & 0.7556 & 0.7756 & 0.8100 & 0.9689 & 0.7978 \\
RefrigerationDevices         & 0.5045 & 0.5120 & 0.5120 & 0.4661 & 0.4757 & 0.4693 \\
Rock                         & 0.7000 & 0.4080 & 0.6440 & 0.5160 & 0.8200 & 0.7600 \\
ScreenType                   & 0.4000 & 0.4053 & 0.3957 & 0.4821 & 0.4624 & 0.4880 \\
SemgHandGenderCh2            & 0.7123 & 0.7750 & 0.6487 & 0.7407 & 0.6627 & 0.6930 \\
SemgHandMovementCh2          & 0.4898 & 0.4618 & 0.3844 & 0.3707 & 0.3947 & 0.3862 \\
SemgHandSubjectCh2           & 0.5387 & 0.5164 & 0.4236 & 0.4102 & 0.4387 & 0.4227 \\
ShakeGestureWiimoteZ         & 0.4320 & 0.5800 & 0.5000 & 0.4360 & 0.5760 & 0.5200 \\
ShapeletSim                  & 0.5289 & 0.6056 & 0.5156 & 0.4633 & 0.5633 & 0.5333 \\
ShapesAll                    & 0.6377 & 0.6500 & 0.5817 & 0.6113 & 0.6583 & 0.6617 \\
SmallKitchenAppliances       & 0.7184 & 0.7189 & 0.6661 & 0.6864 & 0.6757 & 0.6795 \\
SmoothSubspace               & 0.6307 & 0.6907 & 0.6440 & 0.7573 & 0.6160 & 0.6107 \\
SonyAIBORobotSurface1        & 0.8220 & 0.7864 & 0.4842 & 0.7238 & 0.6572 & 0.6998 \\
SonyAIBORobotSurface2        & 0.7356 & 0.8520 & 0.7356 & 0.7051 & 0.8126 & 0.7859 \\
StarLightCurves              & 0.9476 & 0.9520 & 0.9157 & 0.9377 & 0.9253 & 0.9323 \\
Strawberry                   & 0.8811 & 0.9703 & 0.9097 & 0.9514 & 0.8184 & 0.8865 \\
SwedishLeaf                  & 0.7920 & 0.7178 & 0.7568 & 0.7584 & 0.7251 & 0.7702 \\
Symbols                      & 0.6563 & 0.8462 & 0.8318 & 0.8804 & 0.8804 & 0.7337 \\
SyntheticControl             & 0.7973 & 0.7153 & 0.7513 & 0.7793 & 0.6973 & 0.7187 \\
ToeSegmentation1             & 0.5763 & 0.6474 & 0.6149 & 0.6640 & 0.6263 & 0.6184 \\
ToeSegmentation2             & 0.7708 & 0.6385 & 0.6662 & 0.7569 & 0.6708 & 0.5462 \\
Trace                        & 0.5560 & 0.6900 & 0.6100 & 0.6920 & 0.7140 & 0.4720 \\
TwoLeadECG                   & 0.7152 & 0.7638 & 0.9284 & 0.6795 & 0.6681 & 0.5877 \\
TwoPatterns                  & 0.3327 & 0.3689 & 0.3858 & 0.4079 & 0.5154 & 0.4002 \\
UMD                          & 0.6681 & 0.6306 & 0.5528 & 0.4389 & 0.5722 & 0.6569 \\
UWaveGestureLibraryAll       & 0.7267 & 0.7446 & 0.6524 & 0.6744 & 0.6887 & 0.5915 \\
UWaveGestureLibraryX         & 0.5177 & 0.6424 & 0.5869 & 0.5728 & 0.6091 & 0.5824 \\
UWaveGestureLibraryY         & 0.4784 & 0.5650 & 0.5260 & 0.5224 & 0.5176 & 0.5221 \\
UWaveGestureLibraryZ         & 0.5034 & 0.6013 & 0.5503 & 0.5281 & 0.5181 & 0.4954 \\
Wafer                        & 0.9932 & 0.9914 & 0.9737 & 0.9724 & 0.9867 & 0.9950 \\
Wine                         & 0.5963 & 0.6222 & 0.7556 & 0.6296 & 0.4259 & 0.4963 \\
WordSynonyms                 & 0.4122 & 0.4245 & 0.4056 & 0.4088 & 0.4545 & 0.4580 \\
Worms                        & 0.6597 & 0.5922 & 0.6390 & 0.6857 & 0.5896 & 0.6442 \\
WormsTwoClass                & 0.6961 & 0.6779 & 0.7299 & 0.7299 & 0.6649 & 0.6883 \\
Yoga                         & 0.7571 & 0.7417 & 0.6579 & 0.7353 & 0.7488 & 0.7919 \\
\bottomrule
\end{tabular}}
\end{center}

\end{table*}

Table~\ref{tab:IndWindow} shows the accuracy of trained models on individual of a 6-window GeoStat representation. In this instance, one iteration of Laplacian smoothing was used. It is interesting to see that, despite the restriction
\subsubsection{Parameter ablation}
\begin{table*}[]
\caption{Tables showing ablation results for both geometric features and types of statistics used on the UCR 2018 Repository. All categories listed are those delineated by the repository, and all results are given in percent difference in mean classification accuracy. The number of datasets in each category is given in parentheses.}
\label{tab:AblationUCR}

\begin{minipage}{\textwidth}
\begin{center}
\resizebox{0.9\textwidth}{!}{
\begin{tabular}{llllllll}
\toprule
Type             & Position & Velocity & Acceleration & Absolute Curvature & Curvature & Both Curvatures & 2nd Derivatives \\
\midrule
Device (8)       & -6.6283  & -1.9901  & -0.4986 & 0.1156  & -0.2314 & -1.2108 & -2.5083 \\
ECG (6)          & -6.5011  & -1.3558  & 0.6817  & 0.7456  & 0.8118  & 1.2130  & 0.6745  \\
EOG (2)          & -25.9504 & -10.5079 & 8.4299  & 4.6013  & 2.2545  & 8.7351  & 22.2034 \\
EPG (2)          & -1.2097  & -0.3219  & 0.0000  & -0.0404 & 0.0000  & -0.2018 & -0.3632 \\
Hemodynamics (3) & -18.3701 & -1.8471  & 4.6612  & 0.5286  & 5.2521  & 2.7055  & -5.3443 \\
HRM (1)          & -16.3793 & -12.0690 & 0.8621  & 1.7241  & 6.0345  & 21.5517 & 18.1034 \\
Image (32)       & -6.4021  & -2.4677  & 0.5646  & 0.3708  & 0.5704  & 0.6699  & -2.4795 \\
Motion (17)      & -4.6946  & -6.1551  & 1.1760  & 1.2263  & 0.7387  & 1.0527  & 0.5313  \\
Power (1)        & -9.2154  & -1.8680  & -1.1208 & 0.6227  & 0.8717  & -0.3736 & 0.2491  \\
Sensor (30)      & -9.2032  & -3.5468  & -0.2810 & 0.6816  & 0.1681  & 0.6102  & -1.0340 \\
Simulated (8)    & -5.9837  & -11.9659 & -0.9676 & -0.2891 & -0.4647 & -0.7120 & -1.1466 \\
Spectro (8)      & -11.4281 & -4.8479  & 1.7865  & 1.2144  & 1.9862  & 4.3437  & 3.9833  \\
Spectrum (4)     & -3.4562  & 0.0684   & -3.7223 & 2.3696  & 1.1979  & 1.9395  & 5.5012  \\
Traffic (2)      & -4.4758  & 0.1535   & -1.8042 & 3.4192  & 2.5461  & 7.3729  & 10.2489 \\
Trajectory (3)   & -19.3763 & -26.2705 & -3.0396 & -1.8606 & 0.0026  & -4.8231 & -0.1616 \\
\bottomrule \\
\hfill
\end{tabular}}
\end{center}
\begin{center}
\resizebox{0.9\textwidth}{!}{
\begin{tabular}{lllllllll}
\toprule
Type             & Range    & Mean     & Standard Dev.      & Skew     & Kurtosis     & Low Quantiles    & Mid Quantiles     & High Quantiles    \\
\midrule
Device (8)       & 0.1392  & 0.3857  & -0.0477 & -1.3060 & -1.0224 & -0.9925 & -0.3265 & -0.5970  \\
ECG (6)          & -1.3665 & 0.6297  & -0.0333 & 0.2126  & 0.3076  & -0.3893 & -0.2882 & -2.9651  \\
EOG (2)          & 1.1941  & 0.8942  & -0.0747 & -6.9520 & 3.3519  & 2.0084  & 0.5363  & 6.2002   \\
EPG (2)          & 0.0000  & -0.0402 & 0.0000  & 0.0000  & -0.0404 & -0.2815 & 0.0404  & -0.4843  \\
Hemodynamics (3) & -0.9417 & 9.6946  & -0.9543 & -3.6998 & -1.6135 & -1.3052 & -1.0642 & -5.2498  \\
HRM (1)          & 2.5862  & 5.1724  & 1.7241  & -3.4483 & -2.5862 & -2.5862 & -1.7241 & -20.6897 \\
Image (32)       & 0.0032  & -0.1809 & 0.0252  & 0.5852  & -0.2947 & -0.5708 & -1.2744 & -0.6648  \\
Motion (17)      & -0.2531 & 0.0215  & 0.8291  & -0.3244 & 0.0116  & 1.1145  & 1.0723  & -1.8961  \\
Power (1)        & -0.6227 & 0.0000  & -0.7472 & 0.2491  & -0.2491 & -2.7397 & 0.0000  & -1.1208  \\
Sensor (30)      & -0.1480 & -0.5348 & -0.4493 & -0.5928 & -0.5939 & -1.3257 & -0.9714 & -0.4599  \\
Simulated (8)    & -0.3423 & -0.2726 & -0.0676 & -0.3923 & 0.2106  & -0.6146 & -0.8238 & -0.9734  \\
Spectro (8)      & 0.2273  & 1.8783  & 0.0441  & -1.7729 & -0.2518 & -0.7108 & -0.4232 & -3.0456  \\
Spectrum (4)     & 0.5050  & -0.3957 & 0.3904  & 1.0963  & 2.0436  & -2.6752 & 0.2037  & -2.5931  \\
Traffic (2)      & -0.2102 & 0.6328  & 0.2574  & 1.2277  & 1.6080  & -0.4095 & 0.1539  & -0.0939  \\
Trajectory (3)   & -1.7541 & -1.8419 & -1.6562 & 2.6862  & -1.7937 & -6.7577 & -1.9549 & -1.4024  \\
\bottomrule
\end{tabular}}
\end{center}
\strut\end{minipage}

\end{table*}
Table~\ref{tab:AblationUCR} lists ablation results for each of the categories in the UCR 2018 repository. All results are reported in percent difference of average classification accuracy for a KNN model trained on 1-window GeoStat representations with one iteration of Laplacian smoothing. Results for an SVM model with the same parameters are given in the Supplementary Material. Note that choice of number of windows may affect this, but we did not address this directly. Though some numbers are quite large in magnitude, it is important to note that some categories have few datasets and thus may not be representative of all problems. As noted in the main text, both position and velocity information, as well as those of extremal quantiles, appear vital to the success of our representation. The former are not surprising, while the latter presents an interesting direction for future work. The affect of the other quantities, however, appear to be dataset dependent.

\begin{table*}[]
\caption{Ablation results for a fixed random seed on the GFW vessel dataset}
\label{tab:Ablation}
\begin{center}

\begin{minipage}{0.45\textwidth}
\resizebox{0.9\textwidth}{!}{
\begin{tabular}{@{}lll@{}}
\toprule
Removed Features     & KNN    & Percent Change \\ \midrule
None                    & 0.6622 & 0.0000         \\
Active Times           & 0.6793 & 2.5825         \\
Inactive Times         & 0.6685 & 0.9511         \\
Unknown Times          & 0.6820 & 2.9956         \\
Location               & 0.6504 & -1.7785        \\
Speed                  & 0.6278 & -5.1897        \\
Acceleration           & 0.6586 & -0.5491        \\
Smoothed Curvature     & 0.6892 & 4.0840         \\
Rough Curvature        & 0.6847 & 3.4000         \\
All Curvatures         & 0.6847 & 3.4062         \\
All Second Derivatives & 0.6811 & 2.8521         \\
Distance to Port       & 0.6703 & 1.2232         \\
Distance to Shore      & 0.6739 & 1.7649         \\ \midrule
Range                  & 0.6793 & 2.5899         \\
Mean                   & 0.6676 & 0.8101         \\
Std                    & 0.6784 & 2.4439         \\
Skew                   & 0.6775 & 2.3104         \\
Kurtosis               & 0.6793 & 2.5862         \\
Low Quantiles            & 0.6567 & -0.8324         \\
Mid Quantiles            & 0.6450 & -2.6023        \\
High Quantiles           & 0.6929 & 4.6356         \\ \bottomrule
\end{tabular}}
\end{minipage}
\begin{minipage}{0.45\textwidth}
\resizebox{0.9\textwidth}{!}{
\begin{tabular}{@{}lll@{}}
\toprule
Removed Features     & SVM    & Percent Change \\ \midrule
None                    & 0.6956 & 0.0000         \\
Active Times           & 0.6830 & -1.8202        \\
Inactive Times         & 0.7028 & 1.0349         \\
Unknown Times          & 0.6865 & -1.3057        \\
Location               & 0.6532 & -6.1011        \\
Speed                  & 0.6486 & -6.7545        \\
Acceleration           & 0.6883 & -1.0478        \\
Smoothed Curvature     & 0.6983 & 0.3826         \\
Rough Curvature        & 0.7001 & 0.6440         \\
All Curvatures         & 0.6757 & -2.8622        \\
All Second Derivatives & 0.6766 & -2.7374        \\
Distance to Port       & 0.6848 & -1.5588        \\
Distance to Shore      & 0.6848 & -1.5588        \\ \midrule
Range                  & 0.7010 & 0.7782         \\
Mean                   & 0.6956 & -0.0059        \\
Std                    & 0.6974 & 0.2578         \\
Skew                   & 0.6983 & 0.3874         \\
Kurtosis               & 0.6992 & 0.5145         \\
Low Quantiles           & 0.6739 & -3.1235        \\
Mid Quantiles            & 0.6549 & -5.8550        \\
High Quantiles           & 0.6974 & 0.02567        \\ \bottomrule
\end{tabular}}
\end{minipage}
\end{center}

\end{table*}
For the multivariate setting, Table~\ref{tab:Ablation} shows the average effect of the removal of different sets of parameters on the accuracy of both KNN and SVM classifiers for the features given in the GFW dataset. The average is taken over thirty iterations of the 10-fold nested cross-validation procedure. By low, high, and mid quantiles, we mean the lowest four, highest four, and middle three considered of those previously mentioned. The removal of both location (position), speed, and the middle quantiles negatively effect both classifiers. This is unsurprising as the information contained within these (the value of the time series, the rate of change of the time series, and the typical behavior of all quantities) is important basic information.

\end{document}